\pdfoutput=1

\documentclass[11pt]{article}

\usepackage[]{ACL2023}

\usepackage{times}
\usepackage{latexsym}

\usepackage[T1]{fontenc}

\usepackage[utf8]{inputenc}

\usepackage{microtype}

\usepackage{enumitem}
\usepackage{graphicx}
\usepackage{xcolor,colortbl}
\usepackage{amsmath}
\usepackage{booktabs}
\usepackage{multirow}
\usepackage{array}
\usepackage{comment}
\usepackage{bm}
\usepackage{multirow}
\usepackage{scalerel}
\usepackage{textcomp}
\usepackage{amssymb}
\usepackage{pifont}
\usepackage{color,soul}

\usepackage{array}
\newcolumntype{P}[1]{>{\centering\arraybackslash}p{#1}}
\newcolumntype{M}[1]{>{\centering\arraybackslash}m{#1}}

\newcommand{\af}[1]{\textcolor{blue}{#1}}

\newcommand{\STAB}[1]{\begin{tabular}{@{}c@{}}#1\end{tabular}}
\newcommand\redtable[1]{\underline{#1}}

\def\dancer{\scalerel*{\includegraphics{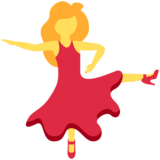}}{\textrm{\textbigcircle}}}
\def\VALSE{VALSE\dancer}
\def\check{\scalerel*{\includegraphics{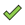}}{\textrm{\textbigcircle}}}
\def\blitz{\scalerel*{\includegraphics{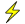}}{\textrm{\textbigcircle}}}

\usepackage{tikz} 
\usetikzlibrary{fit}
\usetikzlibrary{shapes}
\newcommand{\enumber}[1]{%
    \begin{tikzpicture}[remember picture]
        \node[inner sep=0pt](a){#1};
    \end{tikzpicture}%
    \begin{tikzpicture}[overlay, remember picture]
        \node[draw, gray, fit=(a), ellipse, inner sep=.8pt, line width=.5pt]{};
    \end{tikzpicture}%
    }

\definecolor{nicegreen}{HTML}{009B55}
\definecolor{niceorange}{HTML}{F86624}
\definecolor{niceblue}{HTML}{072AC8}

\definecolor{lightyellow}{HTML}{fff2cc}
\definecolor{lightcyan}{HTML}{c9daf8}
\definecolor{lightgreen}{HTML}{d9ead3}
\definecolor{lightpurple}{HTML}{d9d2e9}

\title{MM-SHAP: A Performance-agnostic Metric for Measuring \\ Multimodal Contributions in Vision and Language Models \& Tasks}

\author{Letitia Parcalabescu \and Anette Frank \\
        Computational Linguistics Department \\ Heidelberg University}

\begin{document}
\maketitle
\begin{abstract}
Vision and language models (VL) are known to exploit unrobust indicators in individual modalities (e.g., introduced by distributional biases) instead of focusing on relevant information in each modality. That a unimodal model achieves similar accuracy on a VL task to a multimodal one, indicates that so-called unimodal collapse occurred. However, accuracy-based tests fail to detect e.g., when the model prediction is wrong, while the model used relevant information from a modality. Instead, we propose MM-SHAP, a performance-agnostic multimodality score based on Shapley values that reliably quantifies in which proportions a multimodal model uses individual modalities. We apply MM-\-SHAP in two ways: (1) to compare models for their average degree of multimodality, and (2) to measure for individual models the contribution of individual modalities for different tasks and datasets. Experiments with six VL models – LXMERT, CLIP and four ALBEF variants – on four VL tasks highlight that unimodal collapse can occur to different degrees and in different directions, contradicting the wide-spread assumption that unimodal collapse is one-sided. Based on our results, we recommend MM-SHAP for analysing multimodal tasks, to diagnose and guide progress towards multimodal integration. Code available at \url{https://github.com/Heidelberg-NLP/MM-SHAP}.
\end{abstract}


\begin{figure}[t!]\centering
    \includegraphics[width=\linewidth]{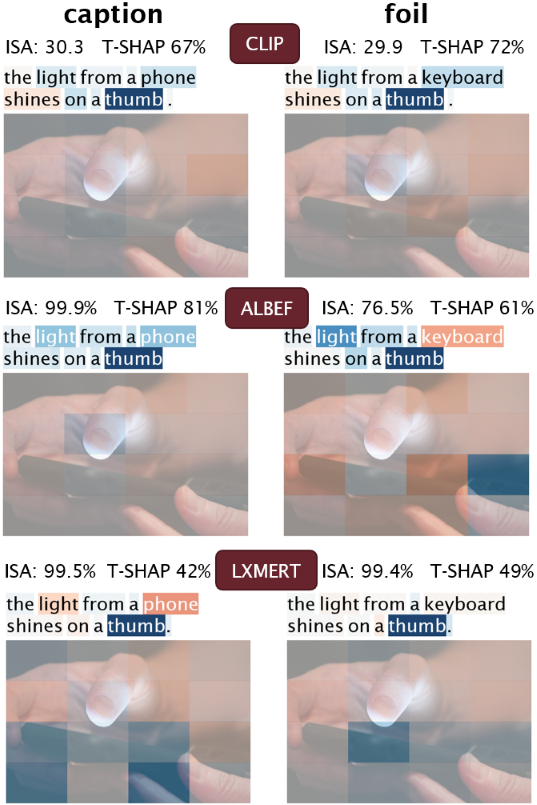}
    \caption{We display image-sentence alignment scores (ISA) and the \emph{textual degree} \texttt{T-SHAP} that measures how much models focus on text rather than the image (with $100 - \texttt{T-SHAP}$\% the corresponding \emph{visual degree})
    for 3 VL models.
    Blue/red highlights on text tokens and image tokens (patches) contribute towards higher/lower ISA.
    Note: CLIP's ISA 
    is an absolute score, while ALBEF and LXMERT predict ISA probabilities.
    See
    Section \ref{sec:exps-results} for more details on
    this figure; App. \ref{app:samples} for more detailed analysis of this instance and more samples.     }
\label{fig:illustrated-contributions}
\end{figure}

\section{Introduction}
Vision and language (VL) tasks are dominated by general-purpose pretrained transformer-based VL models \cite{lu2019vilbert,tan-bansal-2019-lxmert,li2019visualbert,chen2020uniter,Li-etal-2020unicodervl,li2021align-albef}.
But we are only starting to understand why these multimodal (MM) models work so well, and how they utilise and fuse image and text modalities \cite{hessel-lee-2020-multimodal, cao2020behind}. Even worse, these highly parametrised neural VL models, pretrained on large amounts of data, tend to exploit artefacts and statistical correlations in the data \citep{shekhar-etal-2019-evaluating,10.3389/frai.2019.00028}, showing little to no evidence of detailed linguistic or visual understanding \cite{milewski-etal-2022-finding, parcalabescu-etal-2022-valse, thrush2022winoground}.
Statistical biases towards indicators in one modality -- to the detriment of others --
can cause \emph{unimodal collapse} \cite{parcalabescu-etal-2022-valse}, where seemingly MM models exploit one modality that exhibits  
biases, meaning that the MM system effectively reduces to a unimodal model \cite{madhyastha-etal-2018-defoiling}  -- e.g., if a model answers ``How many...?'' questions with ``two'' -- the most frequent answer seen in training
\cite{goyal2017making}.
Unimodal collapse is severe, as it leads to loss of system reliability. It also shows that \textit{multimodal
fusion} is far from being solved. Hence the importance of \textit{measuring mul\-ti\-mo\-dal degree} – the degree to which modalities are used in model predictions – with \textit{reliable metrics}.

To test for unimodal collapse, research has so far focused on performance tests: a VL model is evaluated on a MM task, but one modality crucial for solving it correctly 
is missing,
corrupted \cite{shekhar-etal-2017-foil} or permuted \cite{gat2021perceptualscore}. These tests are indicative of unimodal collapse, but we argue that they are not appropriate to reliably measure the  
contribution of each modality.
Clearly, 
 accuracy reflects whether a model prediction is (in)correct, but 
 it may detect
 illicit
 cases where the model 
prediction is \textit{wrong}, although it 
\textit{does} use crucial 
indicators in a given
modality. Conversely, a prediction might be \textit{correct}, but may be derived from 
unrobust
indicators.
Fig.\ \ref{fig:illustrated-contributions} 
shows very different SHAP-based \textit{contribution patterns} of image regions and text tokens 
leading to model responses of
different image-sentence alignment (ISA) scores (e.g., ALBEF caption vs. foil), 
while yielding same ISA accuracy since
both scores surpass the 0.5 classification threshold.

As an alternative to 
accuracy-based methods, we propose MM-\-SHAP, a \emph{performance-agnostic metric} to quan\-t\-ify and interpret the contribution of individual modalities in VL models.
MM-SHAP is based on Shapley values \cite{Shapley1953}, a theoretically well-\-founded interpretability method from cooperative game theory. 
We apply MM-SHAP to 
quantify
the contribution of specific parts of the input towards 
model predictions.

Our main contributions are:
\begin{enumerate}[label=\roman*), noitemsep]
    \item We propose MM-SHAP,
    a performance-agno\-stic metric to measure the degree of contribution of each modality in VL
    (but not limited to V\&L),
    to \textit{measure the degree to which individual modalities contribute to MM model predictions}.
    We combine MM-SHAP with model accuracy to analyse the degree to which each modality contributes
    to model predictions.
    \item We use MM-SHAP to 1) compare models in terms of their reliance on
different modalities, 2) compare the relevance of different modalities for a given task and dataset, and to 3) zoom in at sample-level to determine the contribution of each modality and each token in each modality for a model prediction (Fig. \ref{fig:illustrated-contributions}).
    \item We conduct experiments with six VL models: 
    LXMERT, 
CLIP 
and four ALBEF 
variants -- on four VL tasks: image-sentence alignment, VQA, 
GQA  
and on the more fine-grained \VALSE{} VL benchmark.  
    \item We identify VL models that are balanced in their usage of two modalities (CLIP), models that show a higher visual degree (LXMERT) or a stronger textual degree (ALBEF).
    \item We show that 1) fine-tuning a model can affect its MM degree and that 2) current VL models do not all collapse towards the same modality, as reported in recent work \cite{frank-etal-2021-vision, gat2021perceptualscore}, but that directions can differ from model to model.

\end{enumerate}

\section{Related Work}
\paragraph{Testing for unimodal collapse}
Strong prediction indicators 
in either modality can cause MM models to ignore weaker indicators in another modality.
Prior work has proposed ways to identify (and remove) such biases from data  \cite{goyal2017making}.



\textit{Foiling approaches} introduce mistakes in image descriptions and test whether VL models notice the discrepancy between image and captions \cite{shekhar-etal-2019-evaluating, parcalabescu-etal-2022-valse}, finding that models are surprisingly insensitive to such foils.  \citet{gat2021perceptualscore}, in a 
similar vein,
exchange 
images with other images or 
captions with other
captions, expecting
that 
inputs with
misleading information in one modality 
incur a decrease in model accuracy.
They 
use an observed
\textit{decrease in task accuracy} to calculate a \textit{perceptual score} as a
measure of the MM degree of a model.
Their findings suggest that across their tested VL models, textual input consistently matters more than visual input.

\textit{Ablation methods} remove information from either 
modality and test whether the model can 
still solve the task. Here,  \citet{frank-etal-2021-vision} 
find that the visual modality matters more than text: VL models suffer from image parts removal
when predicting masked text, but can 
predict masked visual inputs when text input is ablated. This contradicts \citet{gat2021perceptualscore}'s finding, but
their 
investigations have only a single model in common, namely LXMERT.

Hence,
the literature 
agrees that VL models are not as cross-modal as expected -- but 
disagree on whether models rely
more on the textual \cite{gat2021perceptualscore} or on 
the visual modality
\cite{frank-etal-2021-vision}. We argue that a reason for this discrepancy
is that prior work computes MM scores based on model performance.
In our work we argue that methods for measuring a model's MM degree should not rely on accuracy (see §\ref{subsec:why-not-acc} for motivation). Instead, we propose an \textit{accuracy-agnostic} method to measure
the MM degree of VL models, using the \textit{SHAP} \cite{lundberg2017unifiedSHAP} interpretability method that is theoretically suitable to define an MM score.

\paragraph{Interpretability}
Methods for explaining predictions of neural models can be classified into two categories:
\textit{White-box methods}, which require access to specific components of neural architectures and \textit{black-box methods}, which are model-agnostic, requiring only access to model inputs and outputs.

Notable \textit{white-box methods} are: Attention-based methods, which correlate high attention weights with high feature importance. But the equivalence of importance score and attention is debated and must be considered with care \cite{jain-wallace-2019-attention, wiegreffe-pinter-2019-attention} (see App.~\ref{app:against-attention} for a detailed discussion on why attention is inappropriate for
defining an MM score).
Layer-wise relevance propagation \cite{binder2016layer} or gradient-based methods e.g., Grad-CAM \cite{selvaraju2017grad} can also be used to determine
the importance of inputs, but can be deceived by small changes in inputs (adversarial attacks).

Notable \textit{black-box methods} are:
LIME \cite{ribeiro-etal-2016-trust} and its multimodal adaptation DIME \cite{lyu2022dime} approximate the vicinity of the input with a linear function that is interpretable. But depending on the choice of the size of the vicinity, LIME can lead to very disparate results. Methods like RISE \cite{RISE2018} and  SHAP \cite{lundberg2017unifiedSHAP} compute importance scores by randomly masking parts of the input and determining the effect this has on the output. SHAP exhibits great theoretical properties that enable us to define a MM score, as 
we will 
motivate in §\ref{subsec:why-shap}.


\section{Quantifying Multimodal Contributions}
\subsection{
A case for a performance-agnostic 
score
} \label{subsec:why-not-acc}
As a community, we are interested in improving model performance, and thus need to evaluate models using performance metrics such as accuracy. But
in this work we address a complementary question that is only indirectly related to performance. We aim to measure
\textit{how much a given modality matters for model predictions}. 
This is im\-por\-tant for model developers to know, to detect \textit{unimodal collapse}, and to find ways of preventing it.

To date, research tried to measure MM contributions based on accuracy. 
\citet{gat2021perceptualscore} and \citet{frank-etal-2021-vision}, e.g., rely on the difference between a model’s accuracy with and without information from a modality, e.g., to define the \textit{importance of vision} as $V$ = $Acc(vision, text) - Acc(\emptyset, text)$.
This score works well if a MM model shows good performance, but is problematic for wrong model predictions,
since in such cases $Acc(vision, text)$ = $0$, and we expect 
$Acc(\emptyset, text)$ = $0$ too, resulting in 
$V$ = $0$ (or another low value).
But this does not necessarily reflect reality: 
The model may well have relied on the visual modality, but incorrectly.


Even worse, \emph{accuracy-based} methods that completely \emph{delete} \cite{madhyastha-etal-2018-defoiling} or \emph{exchange} \cite{gat2021perceptualscore} information in one modality are ill-defined for image-sentence alignment (ISA): ISA asks a model 
to assess how well two
modalities align, with the rationale that alignment is given if the given modalities (e.g., image and text) contain relevant information that indicates alignment by 'being about the same things or facts'. In case the information conveyed in two modalities is not about the same (type of) things 
(e.g., a picture of a dog paired with a caption talking about 
a cat), the 
modalities do not align.
However, metrics that measure the \emph{importance of vision V} by the impact of deleting it, as $V=Acc(vision, text) - Acc(\emptyset, text)$, are ill-defined for \emph{unaligned} image-sentence pairs: 
A model that uses both modalities to 
correctly predict \emph{misalignment} ($Acc(vision, text)=1$), will also predict a mismatch when the visual information is deleted or exchanged, yielding $Acc(\emptyset, text)=1$. This results in $V=0$, signalling that
no visual 
importance is measured, which is ill-founded in this case.
Hence, accuracy-based scores that rely on deletion of single modalities
are unable to measure multimodal degree on ISA -- an important pretraining task for VL models -- or on zero-shot ISA benchmark tasks such as \VALSE{} \cite{parcalabescu-etal-2022-valse}.

We argue for using \textit{accuracy-agnostic methods} to measure a model's \textit{multimodal degree} and pro\-pose \textit{MM-SHAP,} a metric
that avoids the 
pitfalls of performance-based metrics. We move from $Acc(vision, text)$ to measuring 
the \textit{relative contribution} of vision and text by measuring 
$Contribution(vision,text)$
for a given 
model prediction. We compute the $Contribution$ function using Shapley values, which quantify
a token's contribution
to a model prediction, independently of whether the prediction is correct. 
Importantly, our performance-agnostic way of measuring a model's MM degree in terms of contributions of tokens -- within or across modalities -- will make it possible to clearly separate accuracy-based performance analysis from the study of relative contributions of modalities in MM systems.
This allows us to measure MM
degree in situations where accuracy cannot:
e.g.,\ when model accuracy is low -- as 
in out-of-domain or zero-shot settings.



\subsection{Background on Shapley Values} \label{sec:shap-backgr}
Shapley values\footnote{We refer to \citet{molnar2022} for a gentle introduction into Shapley Values.}
were first introduced in a game theoretical setting to estimate fair rewards among cooperative players \cite{Shapley1953}. For machine learning, the outcome of a game is the model's prediction, the players are parts of the input and are assigned Shapley values that represent the importance of each player \cite{lundberg2017unifiedSHAP}.

We compute Shapley values for pretrained trans\-former-\-based VL models at prediction time. Their input consists of $n$ input tokens (image and text tokens alike). We create subsets $S \subseteq\{1, \ldots, n\}$ of tokens forming a coalition towards the model prediction $val(S)$. Tokens not being part of the subset are masked. $val(\emptyset)$ is the output of the model when all tokens are masked. The Shapley value for a token $j$ follows  formula \eqref{eq:shapley}:
\begin{equation}\label{eq:shapley}
\phi_{j}=\sum_{S \subseteq\{1, \ldots, n\} \backslash\{j\}} \frac{val(S \cup\{j\})-val(S)}{\gamma}
\end{equation}

\noindent
Here, $\gamma = \frac{(n-1)!}{|S| !(n-|S|-1 \mid) !}$ is the normalising factor that accounts
for all possible combinations of choosing 
subset $S$.
When masking $p$ tokens, the coalition possibilities grow exponentially ($2^p$). We thus
approximate the 
Shapley values with Monte Carlo, by randomly sub-sampling $2p+1$ coalitions.

The Shapley value of a token measures its contribution towards the model prediction (e.g., the probability of image-sentence alignment)
and can be \textbf{positive} (increases the model prediction) or \textbf{negative} (decreases it) or \textbf{zero} (no effect).
Shapley values exhibit four defining properties of a fair payout, which are all beneficial for model interpretability: (1) \textit{Efficiency}:
the contributions of all players sum up to the model outcome; (2) \textit{Symmetry}:
any two players that contribute 
equally
are assigned the same payout; (3) \textit{Dummy}:
a non-contributing part is assigned zero value and (4) \textit{Additivity}, enabling us to simply average the Shapley Values to determine the overall player contributions in a game with combined payouts (e.g., the two halves of a soccer match, or ensembling of decision trees).

Most importantly, Shapley values are not based on model accuracy
or performance, but \textit{solely on the model's input and its prediction}, e.g., 
the probability for an image and a caption to match. This is an important property for our
MM score,
since its objective is to quantify \textit{how much 
inputs of either modality matter for prediction} -- even if the cooperation between (multimodal) inputs is not sufficient to reach 
success, i.e., 
yielding 
the correct outcome.

\subsection{MM-SHAP}\label{sec:mm-shap}
For a pretrained VL transformer with $n_T$ text tokens and $n_I$ image tokens, Eq.\ \ref{eq:text-image-contributions} 
defines the textual contribution $\Phi_{T}$ and the image contribution $\Phi_{I}$ towards a prediction 
as the sum of (absolute) Shapley Values (Eq.\ \ref{eq:shapley}) of all textual resp.\ visual tokens:

\begin{equation}\label{eq:text-image-contributions}
\Phi_{T}=\sum_{j}^{n_T} | \phi_{j}| \quad ; \quad \Phi_{I}=\sum_{j}^{n_I} | \phi_{j}|
\end{equation}

\noindent
We consider the magnitude and not the sign of a token contribution\footnote{Contributions can be positive (increase the model prediction) or negative (decrease it) or zero (no effect), see §\ref{sec:shap-backgr}.},
as 
we are interested in 
measuring 
whether a token is active 
in 
a modality -- irrespective of the direction it pushes the prediction into.
Eq.\ \ref{eq:mm-shap} 
defines MM-SHAP as a \textit{proportion} of modality contributions, allowing us to determine a model's
\textit{textual degree} \texttt{T-SHAP} and its \textit{visual degree} \texttt{V-SHAP}:
\begin{equation}\label{eq:mm-shap}
\texttt{T-SHAP} = \frac{\Phi_{T}}{\Phi_{T}+\Phi_{I}} ; \texttt{V-SHAP} = \frac{\Phi_{I}}{\Phi_{T}+\Phi_{I}}
\end{equation}

\noindent
We can extend MM-SHAP 
to any number of mo\-da\-li\-ties. Here we only use image and text.

When generating coalitions, i.e.,  subsets of tokens from which to compute Shapley Values, we do not 
distinguish
image and text tokens, because 
MM-SHAP aims 
to fairly distribute potential token contributions first and 
to aggregate them
modality-wise in a 2\textsuperscript{nd}
step with
Eq.\ \ref{eq:text-image-contributions}.
To \textbf{mask tokens}, we replace text tokens with the [MASK] token;
for images we set pixel values of image patches to zero.
We ensure similar text and image sequence lengths by using more and smaller patches for longer text, and vice versa -- resulting in 16 image patches for the majority of samples in our data.
See App.\ \ref{app:details}.

\subsection{Why SHAP enables a MM score}\label{subsec:why-shap}
Our aim for
MM-SHAP is to estimate the proportion to which text and vision are used by VL models (x\% visual and y\% textual). Defining 
an MM score is nontrivial, since it should
not be based on accuracy, see §\ref{subsec:why-not-acc}.
An MM score should rely on a measure of how much tokens contribute to the output value computed by the model.  
Most interpretablity methods do not directly answer this question 
of how much models use certain features, but use proxies such as gradients or attention. Moreover,
their explanations cannot be added modality-wise in a meaningful way, to define a relative contribution per modality (Cf. App. \ref{app:against-attention} for a discussion on attention).
Luckily, Shapley values compute fair payouts to players (tokens), depending on their contribution to achieving the total payout (the model's prediction). Their theoretically founded properties -- e.g. fair payout between tokens and modalities, or in-sample and between-sample additivity, as detailed in §\ref{sec:shap-backgr} -- allow us to aggregate intra-modal token-level contributions to compute an MM score.

Grounding our MM score in Shapley values bears further advantages, which we discuss next.

\subsection{Ways of using MM-SHAP}\label{subsec:ways-of-using-mmshap}
\paragraph{Sample-level}
MM-SHAP, being based on the contributions of individual image and text tokens, is a sample-level score (Fig.\ \ref{fig:illustrated-contributions}). 
It enables fine-grained analyses of the relevance of tokens from a single or various 
modalities, for each instance.

\paragraph{Dataset and model level}
We can average sample-level MM-SHAP scores into 
dataset-level scores, thanks to the additivity property of Shapley values. Hence it can help
analyse a model across various
datasets, or compare distinct models on a certain
dataset to gain insights of models, datasets / tasks.

\paragraph{Measuring fine-tuning effects}
An accuracy-based MM score is limited when  model performance on a task is very low, since the 
differences between a model's accuracy with correct  vs.\ permuted inputs are small in such cases
(Cf. §\ref{subsec:why-not-acc}).
Since MM-SHAP is based on actual model predictions and not on model performance,
 we can apply MM-SHAP for models with low performance. 
 E.g., we can
 compare a pretrained model's MM score to a fine-tuned version of it that may have lost general task abilities (thus showing
 low accuracy)
after specialising for another task; or we can measure the effectiveness of targeted interventions in fine-tuning to increase a model's reliance on modalities.

Future work could apply MM-SHAP on models accepting different or a wider range of modalities, for tracing a model's MM-SHAP evolution in pre-training, or on data cleaning, by identifying groups of samples
with very unbalanced MM degree -- especially when the accuracy on those samples is high and the model may rely on unimodal cues.

\section{Multimodal 
Contributions across Models and Datasets}
We use MM-SHAP to study MM contributions for different i) model types, ii) 
datasets and iii) tasks. In doing so we iv) re-eval\-u\-ate prior findings 
on visual vs.\ textual unimodal
collapse and v)
showcase MM-SHAP's abilities for 
interpreting
predictions for individual samples, for
error analysis.

We evaluate pretrained VL models with MM-SHAP and complement 
our analysis by measuring the model's task accuracy.
We compare MM-SHAP to a 50\% \texttt{T-SHAP} -- 50\% \texttt{V-SHAP} baseline and gauge how much the model tends towards the textual or visual modality. We hypothesise that in average, V\&L should contribute equally when the model predicts whether the contents of the modalities are aligned (image-sentence alignment).

We test on matching image-captions,
but also on cases with discrepancies between modalities. We break down our incongruity tests into \textit{high discrepancy} (cases of completely mismatching image-captions, Tab.\ \ref{tab:canonical-tasks-results}), and 
\textit{low discrepancy} (cases where a single  word or phrase incurs a mismatch,
Tab.\ \ref{tab:valse-results-main}).

\subsection{Tasks}

\paragraph{Visual Question Answering (VQA)}
is a 
task where transformer-based
VL models
have consistently increased SOTA
performance.
We use the VQA v2.0 
\citep{goyal2017making} and GQA 
\cite{hudson2019gqa} datasets for our experiments.

\paragraph{Image-sentence alignment (ISA)}
VL models are typically
pretrained on predicting an image-sen\-tence alignment score.
We assess their MM contributions in their ``comfort zone'' by letting them predict the alignment of images and captions, in contrast to misalignment to
random captions.
We test 
on 1,500 samples from the 
MSCOCO validation set 
\cite{Lin-etal:2014:mscoco}, 
and
on 
uncommon 
image-caption pairs composed from 
questions and answers from the 
VQA and GQA validation 
sets.

\paragraph{
ISA on fine-grained 
VL 
phenomena} In 
ISA tasks, 
models are typically
confronted with  highly discrepant
negative samples (non-matching
image--captions).
To evaluate 
VL models 
in a more 
fine-grained manner, we 
examine their MM score on the \VALSE{} benchmark \citep{parcalabescu-etal-2022-valse}, 
where foiled captions were created by altering phrases pertaining to
6 specific linguistic phenomena:
existence, counting, plurality, spatial relations, actions, and coreference,
such that image and foiled caption do not
match.
For completeness,
we also test on noun phrase foils 
as introduced 
in the FOILit! dataset \cite{shekhar-etal-2017-foil}.

\subsection{Models}\label{sec:model-description}
\paragraph{LXMERT} by \citet{tan-bansal-2019-lxmert} 
is a dual-stream transformer that combines V\&L in
early fusion using cross-modal attention layers between image and language encoders. 
It was pretrained on MSCOCO \cite{Lin-etal:2014:mscoco} images and captions, and on VQA v2.0 and GQA
ima\-ges, questions and answers.
Its objectives were
(i) multimodal masked word and obj\-ect prediction, (ii) ISA,
and (iii) VQA objectives. For experiments on ISA, VQA and GQA, we use the corresponding 
heads and 
task-specific checkpoints.\footnote{\url{github.com/huggingface/transformers}}

\paragraph{CLIP} by \citet{CLIPradford2021learning} processes image and text with two separate transformer encoders. The resulting image and text representations are combined in late fusion by cross-product. CLIP was trained for ISA in \textit{low discrepancy mode} on 400M image-text pairs
to predict high scores for paired image-text examples and low scores when image-text samples are not paired in the dataset.
With this simple contrastive learning objective,
CLIP shows zero-shot capabilities in e.g.\ object classification, OCR, or activity recognition \cite{CLIPradford2021learning}. 
In our work,
we test CLIP\footnote{\url{github.com/openai/CLIP}} on ISA and \VALSE{}, using the model's image-text alignment score to assess whether it predicts a higher image-text similarity for correct pairs or for foiled image-caption pairs.

\paragraph{ALBEF} by \citet{ALBEFli2021align} combines vision and language with 
early and late fusion.
As in CLIP, 
trans\-former  image and text encoders 
are trai\-ned to 
map the two modalities to a common space. Cross-modal transformer layers further combine the two with 
(i) MM masked word pre\-dic\-tion and (ii) ISA objectives. It was pretrained on Conceptual Captions \cite{sharma2018conceptualcaptions}, SBU Cap\-tions \cite{ordonez2011im2textsbucaptions}, MSCOCO \cite{Lin-etal:2014:mscoco} and Visual Genome \cite{krishna2017visualgenome}.

To analyse
how the
MM
contributions 
are affected by fine-tuning,
we compare 4 ALBEF\footnote{\url{github.com/salesforce/ALBEF}} models fine-tuned on (1) image retrieval on MSCOCO, (2) image retrieval on Flickr30k \cite{plummer2015flickr30k}, (3) visual grounding on RefCOCO+ \cite{yu2016refcoco} and (4) VQA \cite{goyal2017making}.

\subsection{Metrics} \label{subsec:metrics}

We use  
{\bf accuracy} to assess
model performances, 
and {\bf MM-SHAP} to measure
the proportion to which the different modalities contribute. 

With \textbf{MM-SHAP} (def.\ in §\ref{sec:mm-shap}) we aim to analyse the MM contributions in terms of visual (\texttt{V-SHAP}) 
and textual (\texttt{T-SHAP}) degree.
As in our case of two modalities they are complementary  
($\texttt{V-SHAP} = 100 - \texttt{T-SHAP}$), we only report
\texttt{T-SHAP}  (in \%).
We distinguish $\texttt{T-SHAP}_c$ for textual degree in image-\textit{caption} pairs and $\texttt{T-SHAP}_f$ for image-\textit{foil} pairs. As the results 
are very similar, 
we refer to Table \ref{tab:valse-results-appendix} App.\ \ref{app:detailed-results} for 
$\texttt{T-SHAP}_f$ results.

When evaluating VQA and GQA performance, accuracy measures the proportion of correct ans\-wers given pairs of images and questions.
For ISA, we measure the overall accuracy $acc$ of models to classify foils and captions.
We fan out $acc$ into {\bf caption accuracy} $acc_c$ 
(for correctly predicting matching images and captions) and {\bf foil accuracy} $acc_f$ 
(for correctly predicting mismatching images and foils). 
{\bf Pairwise accuracy} $acc_r$ measures the proportion of samples where the ISA score is higher for a correct image-text pair compared to 
its image-foil counterpart.
$acc_r$ is more permissive than $acc$: 
it does not require the ISA score to surpass a classification threshold (of 0.5), but only that image-foil pairs are ranked lower 
than the ground truth 
pairs.

\subsection{Experiments and Results}\label{sec:exps-results}
We test all VL models from
§\ref{sec:model-description} without further tuning
and assess
their task accuracy and 
MM-SHAP
scores in three settings: i)
for VQA on the VQA and GQA datasets;  for ISA ii) with \textit{high discrepancy} 
image-caption pairs 
(from MSCOCO, VQA, GQA) and iii) with
\textit{low discrepancy} pairs from 
\VALSE{}; 
finally iv) we showcase sample-level analyses using MM-SHAP. Table \ref{tab:canonical-tasks-results} shows results on VQA, GQA and ISA; Table \ref{tab:valse-results-main} for \VALSE{}. MM-SHAP varies between samples with a stdev.\ of 
$\sim$12\% across our experiments.

\begin{table*}[t!]
    \small
    \centering
    \resizebox{\linewidth}{!}{
    \begin{tabular}{!{}r cccc|ccccc ccccc ccccc@{}}
        \toprule
        & \multicolumn{4}{@{}c}{\bf Visual Question Answering} & \multicolumn{15}{@{}c}{\bf Image-sentence alignment} \\
        &
        \multicolumn{2}{c|}{\bf VQA} & \multicolumn{2}{c|}{\bf GQA} & \multicolumn{5}{c|}{\bf MSCOCO} & \multicolumn{5}{c|}{\bf VQA} & \multicolumn{5}{c}{\bf GQA}\\
        {\bf Model} & $acc$ & \multicolumn{1}{c|}{$\texttt{T}$}  & $acc$ & \texttt{T} & $acc_c$ & $acc_f$ & $acc_r$  & $\texttt{T}_c$ & \multicolumn{1}{c|@{}}{$\texttt{T}_f$}
        & $acc_c$ & $acc_f$ & $acc_r$  & $\texttt{T}_c$ & \multicolumn{1}{c|}{$\texttt{T}_f$} & $acc_c$ & $acc_f$ & $acc_r$  & $\texttt{T}_c$ & $\texttt{T}_f$\\ 
        \midrule
        \multicolumn{1}{@{}r}{Random} & 0.0&50.0&0.0&50.0&50.0&50.0&50.0&50.0&50.0&50.0&50.0&50.0&50.0&50.0&50.0&50.0&50.0&50.0&50.0\\
        \midrule
        \multirow{1}{*}{LXMERT} & 72.5&{\sethlcolor{lightcyan}\hl{51.5}}&60.3&57.8&71.8&99.1&99.3&\bf{{\sethlcolor{lightyellow}\hl{35.5}}}&\bf{{\sethlcolor{lightyellow}\hl{62.8}}}&66.6&95.9&95.2&{\sethlcolor{lightcyan}\hl{45.7}}&57.5&41.8&96.5&89.9&47.5&59.8\\
        \multirow{1}{*}{CLIP} & -&-&-&-&-&-&99.5&50.3&52.9&-&-&94.0&48.4&47.6&-&-&83.4&47.0&46.0\\
        \multirow{1}{*}{A mscoco}& -&-&-&-&95.9&99.6&99.8&\bf{{\sethlcolor{lightyellow}\hl{63.4}}}&{\sethlcolor{lightyellow}\hl{54.3}}&28.0&99.9&91.0&60.3&59.2&13.1&99.7&83.6&58.3&57.2\\
        \multirow{1}{*}{A flickr}& -&-&-&-&97.3&99.4&99.7&\bf{61.1}&56.6&42.4&99.2&91.8&\bf{61.3}&60.2&23.4&99.5&84.1&58.7&58.1\\
        \multirow{1}{*}{A refcoco} & -&-&-&-&92.3&99.3&99.7&56.6&58.9&49.8&99.1&90.0&57.8&58.6&25.0&98.4&85.6&58.2&59.3\\
        \multirow{1}{*}{A vqa} & 76.0&66.7&-&-&99.9&0.0&33.4&\bf{64.1}&\bf{62.8}&100.0&0.0&60.2&58.2&60.0&100.0&0.0&52.6&61.7&\bf{62.4}\\
        \bottomrule
    \end{tabular}
    }
    \caption{Task accuracy and MM score on canonical tasks. \texttt{T} is \texttt{T-SHAP} (in \%). $\texttt{V-SHAP} = 100-\texttt{T-SHAP}$.
    $acc_r$ is pairwise ranking accuracy, counting predictions as correct if $p(caption,img) > p(random,img)$.
    {\bf A} stands for ALBEF fine-tuned for different tasks: image retrieval on MSCOCO and Flickr30k; visual grounding on RefCOCO+ and VQA.
    Overall foil task performance is the mean of $acc_c$ and $acc_f$ (equal nb.\ of samples, all pairs). }
    
    \label{tab:canonical-tasks-results}
\end{table*}

\begin{table*}[t!]
    \small
    \centering
    \resizebox{\linewidth}{!}{
    \begin{tabular}{ l r r@{\hskip 0.2in}r@{\hskip 0.2in}rrr@{\hskip 0.2in}r@{\hskip 0.2in}rr@{\hskip 0.2in}rr@{\hskip 0.2in}r@{\hskip 0.2in}c l }
        \toprule
        \multirow{2}{*}{\bf Metric} & \multirow{2}{*}{\bf Model} &
        \multicolumn{1}{c|}{\bf Existence} & \multicolumn{1}{c|}{\bf Plurality} & \multicolumn{3}{c|}{\bf Counting} & \multicolumn{1}{c|}{\bf Sp.rel.$\ddagger$} & \multicolumn{2}{c|}{\bf Action} & \multicolumn{2}{c|}{\bf Coreference} & \multicolumn{1}{c|}{\bf Foil-it!} & \multicolumn{1}{c|}{{\bf Avg.}} &
        \multicolumn{1}{c}{{\bf MM}}\\
        && \multicolumn{1}{c|}{quantifiers} & \multicolumn{1}{c|}{number} & \multicolumn{1}{c}{bal.$\dagger$} & \multicolumn{1}{c}{sns.$\dagger$} & \multicolumn{1}{c|}{adv.$\dagger$} & \multicolumn{1}{c|}{relations} & \multicolumn{1}{c}{repl.$\dagger$} & \multicolumn{1}{c|}{swap$\dagger$} & \multicolumn{1}{c}{std.$\dagger$} & \multicolumn{1}{c|}{clean} &
        \multicolumn{1}{c|}{nouns}& 
        \multicolumn{1}{c|}{$\pm$ stdev.} &
        \multicolumn{1}{c}{skew}\\ 
        \midrule
        & \multicolumn{1}{r}{Random} & 50.0 & 50.0 & 50.0 & 50.0 & 50.0 & 50.0 & 50.0 & 50.0 & 50.0 & 50.0 & 50.0 & 50.0$\pm$0 & \\
        \midrule
        
        \multirow{6}{*}{$acc_r$}
        & \multirow{1}{*}{CLIP} & 66.9 & 56.2 & 62.1 & 62.5 & 57.5 & 64.3 & 75.6 & 68.6 & 52.1 & 49.7 & 88.8 & 64.0$\pm$11\\
        & \multirow{1}{*}{LXMERT} & 78.6 & 64.4 & 62.2 & \enumber{69.2} & \enumber{42.6} & 60.2 & 54.8 & 45.8 & 46.8 & 44.2 & 87.1 & 59.6$\pm$15\\
        & \multirow{1}{*}{A mscoco} & 78.6&80.1&71.8&74.3&68.9&74.6&79.8&62.6&62.2&59.6&{\bf97.0}&{\bf {\sethlcolor{lightpurple}\hl{73.6}}}$\pm$11\\
        & \multirow{1}{*}{A flickr} & 80.6&78.9&71.0&73.6&64.3&73.3&82.4&55.5&59.9&57.7&96.6&72.1$\pm$12\\
        & \multirow{1}{*}{A refcoco} & 73.1&69.0&67.9&\enumber{70.7}&\enumber{45.7}&68.6&79.9&58.9&52.7&43.3&96.5&{\sethlcolor{lightpurple}\hl{66.0}}$\pm$15\\
        & \multirow{1}{*}{A vqa} & 40.8&63.3&49.0&49.2&23.2&61.9&51.7&52.0&55.9&43.3&67.2&{\sethlcolor{lightpurple}\hl{50.7}}$\pm$12\\

        \cmidrule{2-15}
        \multirow{5}{*}{$acc$}
        & \multirow{1}{*}{LXMERT} & 55.8	&55.1	&52.0	&55.4	&49.4	&50.7	&51.1&	48.5	&49.8	&49.0	&70.8	&53.4$\pm$6\\
        & \multirow{1}{*}{A mscoco} & 56.7	&60.2&	55.4	&53.9	&56.0	&52.3	&63.7&	54.0	&52.7	&52.0	&76.3	&57.6$\pm$7\\
        & \multirow{1}{*}{A flickr} & 55.6	&56.3	&53.8	&53.3	&55.4	&52.3	&64.9	&48.9&	50.0	&50.0	&70.5	&55.5$\pm$6\\
        & \multirow{1}{*}{A refcoco} & 53.4	&56.3	&51.1	&51.1	&48.4	&51.1	&63.1&	51.2&	50.7	&49.3	&77.4	&54.8$\pm$8\\
        & \multirow{1}{*}{A vqa} & 52.8	&50.0	&50.0&	50.0	&51.1	&53.5	&50.0	&50.0	&51.4	&50.0	&53.7	&51.1$\pm$1\\

        
        
        \cmidrule{2-15}
        \multirow{6}{*}{\STAB{\rotatebox[origin=c]{60}{$\texttt{T-SHAP}_c$}}}
        & \multirow{1}{*}{CLIP} & 44.7&52.3&51.5&51.8&52.1&50.9&50.0&49.7&52.1&52.6&49.9&{\sethlcolor{lightgreen}\hl{50.7}}$\pm$2 & bal.\\
        & \multirow{1}{*}{LXMERT} & 51.7&{\bf37.1}&46.5&\enumber{47.3}&\enumber{46.4}&{\bf36.6}&42.1&42.2&{\bf38.2}&{\bf37.2}&\redtable{\bf{36.1}}&{\sethlcolor{lightgreen}\hl{41.9}}$\pm$5 & vis.\\
        & \multirow{1}{*}{A mscoco} & 56.7&{\bf63.5}&58.3&58.0&59.5&{\bf64.1}&{\bf61.7}&{\bf61.5}&{\bf61.9}&{\bf61.4}&{\bf63.9}&60.9$\pm$3 & txt.\\
        & \multirow{1}{*}{A flickr} & 59.5&\bf{61.7}&59.6&59.8&59.5&\bf{61.6}&59.8&58.9&60.9&{\bf61.9}&{\bf63.5}&60.6$\pm$1 & txt.\\
        & \multirow{1}{*}{A refcoco} & 53.3&57.2&55.4&\enumber{55.1}&\enumber{55.8}&57.0&54.5&54.4&57.9&58.9&\redtable{56.8}&{\sethlcolor{lightpurple}\hl{56.0}}$\pm$2 & txt.\\
        & \multirow{1}{*}{A vqa} & {\bf64.6}&{\bf63.6}&{\bf62.5}&{\bf61.4}&{\bf63.4}&{\bf63.0}&59.3&60.3&{\bf63.6}&{\bf63.1}&{\bf62.1}&{\bf{\sethlcolor{lightgreen}\hl{62.4}}}$\pm$2 & txt.\\
        

        \bottomrule
    \end{tabular}    }
    \caption{ Performance and MM scores of VL models on the \protect\VALSE{} benchmark. We bold-face high accuracies and multimodally unbalanced models on tasks.
    $acc_r$: the pairwise ranking accuracy,
    considering predictions as correct if $p(caption,img) > p(foil,img)$.
    $acc$: Overall ISA accurracy.
    {\bf A} stands for different fine-tunings of ALBEF: image retrieval on MSCOCO and Flickr30k, visual grounding on RefCOCO+ and VQA.
    $\dagger${\bf bal.} Counting balanced. $\dagger${\bf sns.} Counting small numbers. {\bf adv.} Counting adversarial. {\bf repl.} Action replacement. {\bf swap.} Actant swap. $\ddagger$ {\bf Sp.rel.} Spatial relations. $\dagger${\bf std.} Coreference standard. {\bf MM skew}: Modality on which a model relies more: {\bf bal.} balanced, {\bf vis.} visual, {\bf txt.} textual.  We refer to Table \ref{tab:valse-results-appendix} in App. \ref{app:detailed-results} for more fanned out results.}
    \label{tab:valse-results-main}
\end{table*}
\paragraph{High discrepancy ISA} (Table \ref{tab:canonical-tasks-results}) shows that
$acc_r$ scores for ISA on MSCOCO, VQA, GQA 
are high for all models. This is expected as 
they have been pretrained for ISA -- only ALBEF vqa stands out: it lost its ISA performance by fine-tuning on VQA.
LXMERT has highest $acc_r$ for ISA on VQA and GQA, since for its last 10 epochs it was trained on 
these datasets.

For ISA, we observe the models scattering around the hypothesised 50\% balance for \texttt{T-SHAP}, with CLIP being the most balanced one,
especially on MSCOCO.
This is expected since CLIP is a two-branch model where the two modalities communicate in late fusion, in other words, CLIP keeps all information from the textual and visual branches separate until the very end. By contrast, LXMERT has a low textual degree of only {\sethlcolor{lightyellow}\hl{35.5}}\%, while ALBEF models are more textual.

Given \textit{highly diverging} foil pairs,
$\texttt{T-SHAP}_c$ and $\texttt{T-SHAP}_f$ differ prominently: LXMERT moves from weak 
to higher textual degree ({\sethlcolor{lightyellow}\hl{35.5}}  to {\sethlcolor{lightyellow}\hl{62.8}}\%) and inversely for ALBEF mscoco ({\sethlcolor{lightyellow}\hl{63.4}} to {\sethlcolor{lightyellow}\hl{54.3}}\%).

\paragraph{Canonical VL tasks}
Results on VQA and GQA in Table \ref{tab:canonical-tasks-results} -- with ALBEF fine-tuned for VQA and LXMERT fine-tuned on VQA and GQA\footnote{We do not test CLIP and the other ALBEF models on VQA because they do not have corresponding VQA heads.} -- show high model accuracy. \texttt{T-SHAP} is higher for VQA ({\sethlcolor{lightcyan}\hl{51.5}}\%) than for ISA ({\sethlcolor{lightcyan}\hl{45.7}}\% $acc_c$),
which is interesting, since LXMERT was more visually focused on ISA. It seems like ALBEF vqa's and LXMERT's training on VQA increases the impact of the textual modality to the detriment of the visual one. This aligns with earlier findings that in VQA tasks, linguistic indicators (e.g., ``How many...?'') give away the most likely answer (two) \cite{goyal2017making}.

\paragraph{Low discrepancy ISA}
on \VALSE{} in Table~\ref{tab:valse-results-main}. For $\texttt{T-SHAP}_c$ we bold-face 
high deviations from the 50\% \texttt{T-SHAP} baseline 
(values > 61\% and <40\%). We note
that the scores 
do not deviate much from the baseline. CLIP is the multimodally most balanced model, with an average $\texttt{T-SHAP}_c$ of {\sethlcolor{lightgreen}\hl{50.7}}\% across all instruments, which is expected, as argued for high discrepancy ISA above. By contrast, LXMERT skews towards the visual modality with an average $\texttt{T-SHAP}_c$ of {\sethlcolor{lightgreen}\hl{41.9}}\%, while ALBEF focuses on text -- its variants showing $\texttt{T-SHAP}_c$ values of 56\% to {\sethlcolor{lightgreen}\hl{62}}\%. This is consistent with our results for high discrepancy ISA in Table \ref{tab:canonical-tasks-results}.

\paragraph{Accuracy vs. MM-SHAP}
On \VALSE{},
accuracies do not correlate with MM-SHAP (see App. \ref{app:shap-acc} for details).
This suggests that MM-SHAP \textit{complements accuracy} in
assessing MM contributions.
As 
\citet{parcalabescu-etal-2022-valse} observe, models are better with some instruments (noun phrases, existence) as opposed to  others (actions, coreference). Our work adds 
the multimodal score MM-SHAP as a new dimension of analysis. Some models exhibit strong divergences
in \texttt{T-SHAP} across phenomena:
LXMERT is strongly
visually focused for plurality, spatial relations, noun phrases; 
also ALBEF's 
textual bias is especially strong
for these phenomena.

\paragraph{Model bias}

For overall ISA results on \VALSE{}, Table \ref{tab:valse-results-main} 
shows
that despite
varying model accuracies (stdev.\ for $acc_r$ across phenomena 
$\pm$11-15\%),
MM-SHAP is relatively stable
($\pm$1-5\% stdev.)
even when data distributions differ:
E.g., 
\emph{counting adversarial} 
contains foils in
number ranges 
0 to 3, while for captions 
numbers are higher than 4. The piece serves as a sanity check for
biased models that may prefer the more frequently found small numbers. 
For LXMERT and ALBEF refcoco $acc_r$ drops for counting \emph{small numbers} to counting \emph{adversarial} 
(encircled numbers in Tab.~\ref{tab:valse-results-main}) from 69.2\% to 42.6\% for LXMERT and from 70.7\% to 45.7\% for ALBEF -- while $\texttt{T-SHAP}_c$ stays remarkably constant (47.3\% to 46.4\% and 55.1\% to 55.8\%).
Even for phenomena that suffer from plausibility bias \cite{parcalabescu-etal-2022-valse}, $\texttt{T-SHAP}$ varies little, while accuracies differ. Stable MM-SHAP scores
highlight our MM score's ability to measure how much the input modalities matter for
model predictions -- 
irrespective of their
correctness~--, complementing accuracy.
Further
results in App.\ \ref{app:diff-cap-foils} compare model performances on foils vs. captions, supporting MM-SHAP's stability while accuracy varies.


\paragraph{Fine-tuning effects}
For the four fine-tuned ALBEF models evaluated
on \VALSE{}, we observe that i) the models fine-tuned for image retrieval (mscoco, flickr) are good at predicting ISA ({\sethlcolor{lightpurple}\hl{73.6}}\% $acc_r$ for ALBEF mscoco) but not those
for VQA (ALBEF vqa {\sethlcolor{lightpurple}\hl{50.7}}\%) and referring expressions (ALBEF refcoco {\sethlcolor{lightpurple}\hl{66.0}}\%). This is expected, since ISA and image retrieval are very similar tasks.
Interestingly, not only accuracy, but also the MM score changes, making ALBEF vqa
more focused on text ({\sethlcolor{lightgreen}\hl{62.4}}\% avg.\ $\texttt{T-SHAP}_c$ across VALSE) compared to referring expressions (ALBEF refcoco {\sethlcolor{lightpurple}\hl{56.0}}\%). Notably, MM-SHAP being accuracy-agnostic, we can compute indicative scores even when a fine-tuned model fails the task completely, like ALBEF vqa that always predicts the foil class on captions.

\paragraph{Sample-level analysis}
Fig.\ \ref{fig:illustrated-contributions} shows
ISA pre\-dic\-tions of CLIP, ALBEF mscoco and LXMERT, and their \texttt{T-SHAP} values for caption and foil.
LX\-MERT correctly predicts high ISA between image and \emph{caption} (left), although the regions contributing most 
(in blue) are not all reasonable, since the `phone' token is not correctly grounded. ALBEF mscoco and CLIP also assign very high ISA scores, while using well-justified image regions 
for thumb and phone.
On the \textit{foil} (right), 
LXMERT's contributing tokens change, with the phone region in the image mistakenly contributing to a high ISA. Favourably for ALBEF, the `keyboard' text token contributes towards lowering the ISA, unlike for CLIP and LXMERT, where the `keyboard' token raises the ISA.
For more examples see App.\ \ref{app:samples}. We also showcase how attention does \textit{not} reflect negative impact of tokens on a model's prediction -- which is very important in e.g., assessing the impact of foil words -- in App. \ref{app:attention-no-neg} and Fig. \ref{fig-app:attention-low-discr}.

\subsection{Comparison to other MM metrics}
We can only compare to other MM scores for VQA, 
because accuracy-based MM scores that 
delete information cannot apply to ISA (as argued in §\ref{subsec:why-not-acc}). 

Unsurprisingly LXMERT's accuracy when deleting the image is 31\%; when deleting the text it is 8\%, since excluding  the image
should negatively affect accuracy more than excluding text in VQA, where at least the answer type can be better inferred from 
the text (should be numeral for ``How many''). But this ablation 
tells us more about the task definition than 
a model's reliance on modalities.

The Perceptual Score \cite{gat2021perceptualscore} computes the per-sample difference between the model's accuracy when working with the correct image and text as input and with a random image or text.
LXMERT's Perceptual Score \cite{gat2021perceptualscore} 
is 32.5 visual, 41.6 textual 
(relying more on text), but we argued in §\ref{subsec:why-not-acc} that does not reflect cases where a model makes a wrong prediction because it
failed to interpret the right cues correctly. 
MM-SHAP rates LXMERT vqa as  
balanced 
({\sethlcolor{lightcyan}\hl{51.5}\% \texttt{T-SHAP})}.


\subsection{On the need of a MM
score}

Our experiments show that a models' reliance on a modality can vary with each 
task, dataset and instance.
While prior work found that the models they analysed
\emph{all} 
prefer a \emph{single} modality that they rely on most, our analyses 
show that \emph{different} VL models behave \emph{differently} on \emph{the same} task: 
ALBEF is rather textual, CLIP ba\-lan\-ced, LXMERT shows higher visual 
degree.

For LXMERT, we side with \citet{frank-etal-2021-vision}, who found it to have a higher visual preference -- this aligns with our analysis yielding a  T-SHAP of
41.9\%.
We therefore disagree with \citet{gat2021perceptualscore}, 
who found a preference towards text.

Clearly, we do not assume
that a MM model must rely equally on multiple modalities, but there are cases where unimodal collapse is unwanted, i.e., 
a model gives the right answer for the wrong reason in tasks such as VQA. MM-SHAP helps identify how much models rely on each modality.

\section{Conclusions and Future Work}
We present MM-SHAP, a performance-agnostic metric that measures the MM degree of VL models at dataset and sample level. 
Our results show that \textit{on the same task, dataset}, and on specific \textit{instances}, different types of models rely on modalities to different degrees and in different directions.
Using MM-\-SHAP we are the first to quantify changes in a model's MM degree through fine-tuning.
Our analyses show that de\-grees of MM contributions can be or\-tho\-go\-nal to task performance, supporting the need for performance-agnostic metrics.
MM-SHAP is ap\-pli\-cable to further modalities. It 
enables model-bas\-ed data cleaning and thus, dataset bias removal. Finally, 
it can serve as a diagnostic tool for improving MM fusion methods. 

MM-SHAP can be used for testing true model understanding at a global and at instance level, and whether a model is giving the right answer for the right reasons, at corpus -- and instance-level -- which is not guaranteed for performance-dependent metrics. It can help us track MM contributions during (pre-)\-training and towards assessing and eventually
predicting how much a model needs to rely on how many and which modalities in a given task or instance case -- and how to explain this. We hence believe that many future research questions will profit from our MM 
score as an unbiased MM contribution metric, with 
AI research advancing to include more and more 
modalities beyond vision and language \cite{girdhar2023imagebind}: acoustics, haptics, emotion, and more (cf.\ \citet{parcalabescu-etal-2021-multimodality}).

\section*{Limitations}\label{app:limitations}
This work focused on assessing multimodal degree for recent 
English VL models.
The following limitations can be relevant for future work.

We only evaluated a limited number of models in a zero-shot setting using their image-sentence alignment and VQA heads. Future work might be interested in assessing more models and tracking the evolution of MM-SHAP scores during model pretraining and finetuning.

This work applied MM-SHAP to VL encoders. We leave it for future work to investigate autoregressive (decoder-only) VL models. In the time it took to review and publish this work, we already encountered efforts to apply Shapley Values for interpreting VL models in \citet{cafagna2023interpreting}.

We only applied ML-SHAP to VL models. Future work might be interested in models working with other or additional modalities beyond vision and language.

Computing all possible coalitions between input tokens for Shapley Values is infeasible because their number is exponential in the number of tokens ($2^p$). Therefore we perform Monte Carlo approximation by randomly sub-sampling $2p+1$ coalitions. This results in approximate MM-SHAP scores per sample. We argue that as an alternative, one can simply increase the number of sampled coalitions for more exact measurements (as we did 10-fold for Fig.~\ref{fig:illustrated-contributions} and the examples in Appendix \ref{app:samples}) -- at the cost of increasing the environmental footprint.
But it is not necessary to increase the number of samples when estimating MM-SHAP at dataset level, because the number of coalitions has very little effect on a data-set wide range -- given that approximation fluctuations average out.

To compute MM-SHAP at data-set level, one needs to run models in inference mode $2p+1$ times, where $p$ is the number of tokens to mask (around 40 in average for MSCOCO-sized captions). On an NVIDIA Titan X GPU, computing MM-SHAP for one image-caption pair can take ~2 seconds for ALBEF, ~3 seconds for CLIP. LXMERT is the most expensive and needs ~15 seconds, because it computes image features with a CNN backbone for every masking configuration.

\section*{Ethical Considerations}
This paper uses publicly available datasets and models and therefore could carry on their potential biases \cite{meister2022gender, Garcia_2023_CVPR} and imperfections. However, the method presented in this paper  enables model and dataset interpretation and we hope that it \emph{can help future work locate harmful biases}.

\section*{Acknowledgements}
The authors would like to thank the anonymous reviewers, Albert Gatt and Emanuele Bugliarello for their useful suggestions. Thanks go to Nils Trost for assisting with the visualisations.

The authors acknowledge support by the state of Baden-Württemberg through bwHPC
and the German Research Foundation (DFG) through grant INST 35/1597-1 FUGG.

\bibliography{anthology,refs}
\bibliographystyle{acl_natbib}

\clearpage
\appendix

\section{Experimental Details} \label{app:details}
\paragraph{Masking}
VL models predict their outputs (such as ISA) on full and uncorrupted image and text inputs. To compute Shapley values and with them the MM-SHAP score, we create coalitions by masking image and text tokens.
For \textbf{masking text}, we replace the text tokens with the [MASK] token.

For \textbf{masking images} we mask out image patches setting pixel values to zero. The patches are the regions for which we compute Shapley values, as visualised in Figures \ref{fig:app-version-fig1} to \ref{fig-app:actions-hard}. By masking these patches, the SHAP algorithm can estimate how the prediction of the model changes in all possible combinations of their presence or absence. 

After zero-ing out the patches, the models work as usual:
LXMERT with the Faster-RCNN backbone computes image features and extracts image tokens. Working on the image level has the upside that no neighborhood information can leak into each image token: If we were to mask out on feature-level of the Faster-RCNN, i.e., on rectangular regions, the other regions would possibly “know about” the other regions due to the hierarchical structure of the CNN.
For CLIP, the CLIP image encoder works as usual: It works internally with 32x32 patches of images in which we have already zeroed out information.

Therefore this masking procedure has the upside of being  directly applicable to different types of VL model architectures, since some apply transformers directly on the image (CLIP and ALBEF), while others compute image tokens (features) with a different CNN-based backbone (LXMERT).

For computing Shapley values, we aim for a balance between text and image sequence length to make MM-SHAP adaptable to variable caption lengths and variable image sizes. Therefore we use the text length to dynamically determine patch sizes: For longer text, we use more and smaller patches and for shorter text, less but bigger patches.
In the majority of our experiments, we have 16 image patches. We illustrate the image tiling in the top right of Figures \ref{fig:app-version-fig1} to \ref{fig-app:actions-hard}.

This masking procedure 
has several advantages: i) It adapts to variable caption lengths and variable image sizes, and ii) it directly applies to different types of VL model architectures, since some apply transformers directly on the image (CLIP and ALBEF), while others compute image tokens (features) with a different CNN-based backbone (LXMERT).

\paragraph{Special tokens}
When computing token-wise contributions, we do not take [SEP] and [CLS] tokens into account (i.e., they are always assigned zero contribution), since their functionality is to aggregate cross-modal information, e.g. for classification, and hence they cannot be attributed to one modality exclusively.

\begin{table*}[t!]
    \small
    \centering
    \resizebox{\linewidth}{!}{
    \begin{tabular}{ l r r@{\hskip 0.2in}r@{\hskip 0.2in}rrr@{\hskip 0.2in}r@{\hskip 0.2in}rr@{\hskip 0.2in}rr@{\hskip 0.2in}r@{\hskip 0.2in}c l }
        \toprule
        \multirow{2}{*}{\bf Metric} & \multirow{2}{*}{\bf Model} &
        \multicolumn{1}{c|}{\bf Existence} & \multicolumn{1}{c|}{\bf Plurality} & \multicolumn{3}{c|}{\bf Counting} & \multicolumn{1}{c|}{\bf Sp.rel.$\ddagger$} & \multicolumn{2}{c|}{\bf Action} & \multicolumn{2}{c|}{\bf Coreference} & \multicolumn{1}{c|}{\bf Foil-it!} & \multicolumn{1}{c|}{{\bf Avg.}} &
        \multicolumn{1}{c}{{\bf MM}}\\
        && \multicolumn{1}{c|}{quantifiers} & \multicolumn{1}{c|}{number} & \multicolumn{1}{c}{bal.$\dagger$} & \multicolumn{1}{c}{sns.$\dagger$} & \multicolumn{1}{c|}{adv.$\dagger$} & \multicolumn{1}{c|}{relations} & \multicolumn{1}{c}{repl.$\dagger$} & \multicolumn{1}{c|}{swap$\dagger$} & \multicolumn{1}{c}{std.$\dagger$} & \multicolumn{1}{c|}{clean} &
        \multicolumn{1}{c|}{nouns}& 
        \multicolumn{1}{c|}{$\pm$ stdev.} &
        \multicolumn{1}{c}{skew}\\ 
        \midrule
        & \multicolumn{1}{r}{Random} & 50.0 & 50.0 & 50.0 & 50.0 & 50.0 & 50.0 & 50.0 & 50.0 & 50.0 & 50.0 & 50.0 & 50.0$\pm$0 & \\
        \midrule
        
        \multirow{6}{*}{$acc_r$}
        & \multirow{1}{*}{CLIP} & 66.9 & 56.2 & 62.1 & 62.5 & 57.5 & 64.3 & 75.6 & 68.6 & 52.1 & 49.7 & 88.8 & 64.0$\pm$11\\
        & \multirow{1}{*}{LXMERT} & 78.6 & 64.4 & 62.2 & \enumber{69.2} & \enumber{42.6} & 60.2 & 54.8 & 45.8 & 46.8 & 44.2 & 87.1 & 59.6$\pm$15\\
        & \multirow{1}{*}{A mscoco} & 78.6&80.1&71.8&74.3&68.9&74.6&79.8&62.6&62.2&59.6&{\bf97.0}&{\bf {\sethlcolor{lightpurple}\hl{73.6}}}$\pm$11\\
        & \multirow{1}{*}{A flickr} & 80.6&78.9&71.0&73.6&64.3&73.3&82.4&55.5&59.9&57.7&96.6&72.1$\pm$12\\
        & \multirow{1}{*}{A refcoco} & 73.1&69.0&67.9&\enumber{70.7}&\enumber{45.7}&68.6&79.9&58.9&52.7&43.3&96.5&{\sethlcolor{lightpurple}\hl{66.0}}$\pm$15\\
        & \multirow{1}{*}{A vqa} & 40.8&63.3&49.0&49.2&23.2&61.9&51.7&52.0&55.9&43.3&67.2&{\sethlcolor{lightpurple}\hl{50.7}}$\pm$12\\

        \cmidrule{2-15}
        \multirow{5}{*}{$acc$}
        & \multirow{1}{*}{LXMERT} & 55.8	&55.1	&52.0	&55.4	&49.4	&50.7	&51.1&	48.5	&49.8	&49.0	&70.8	&53.4$\pm$6\\
        & \multirow{1}{*}{A mscoco} & 56.7	&60.2&	55.4	&53.9	&56.0	&52.3	&63.7&	54.0	&52.7	&52.0	&76.3	&57.6$\pm$7\\
        & \multirow{1}{*}{A flickr} & 55.6	&56.3	&53.8	&53.3	&55.4	&52.3	&64.9	&48.9&	50.0	&50.0	&70.5	&55.5$\pm$6\\
        & \multirow{1}{*}{A refcoco} & 53.4	&56.3	&51.1	&51.1	&48.4	&51.1	&63.1&	51.2&	50.7	&49.3	&77.4	&54.8$\pm$8\\
        & \multirow{1}{*}{A vqa} & 52.8	&50.0	&50.0&	50.0	&51.1	&53.5	&50.0	&50.0	&51.4	&50.0	&53.7	&51.1$\pm$1\\

        \cmidrule{2-15}
        \multirow{5}{*}{$acc_c$}
        & \multirow{1}{*}{LXMERT} & 41.6	&68.0	&50.9	&50.0	&61.5	&73.1	&35.8	&36.8	&81.2	&80.8	&72.3	&59.3$\pm$17\\
        & \multirow{1}{*}{A mscoco} & 18.4&93.2&26.7&23.7&34.6&95.9&66.2&64.9&87.0&89.4&96.1&63.3$\pm$32\\
        & \multirow{1}{*}{A flickr} & 28.7&94.0&43.1&41.2&50.8&96.8&65.1&64.2&91.5&96.2&97.5&69.9$\pm$26\\
        & \multirow{1}{*}{A refcoco} & 33.7&89.8&41.8&31.0&57.2&93.1&72.5&75.0&81.4&90.4&92.7&69.0$\pm$24\\
        & \multirow{1}{*}{A vqa} & 0.0&0.0&0.0&0.0&0.0&0.0&0.0&0.0&0.0&0.0&0.0&0.0$\pm$0\\
        
        \cmidrule{2-15}
        \multirow{5}{*}{$acc_f$}
        & \multirow{1}{*}{LXMERT} & 70.1	&42.2	&53.0	&60.8	&37.3	&28.4	&66.4	&60.2	&18.4	&17.3	&69.3	&47.6$\pm$20\\
        & \multirow{1}{*}{A mscoco} & 91.5&27.1&82.0&87.2&80.9&9.2&61.7&42.3&16.1&12.5&52.1&51.1$\pm$32\\
        & \multirow{1}{*}{A flickr} & 82.4&18.5&66.4&70.9&58.6&7.1&63.3&38.8&8.2&4.8&42.4&41.9$\pm$28\\
        & \multirow{1}{*}{A refcoco} & 71.3&19.4&62.0&72.9&41.8&10.5&53.2&29.7&18.4&8.7&61.19&40.8$\pm$25\\
        & \multirow{1}{*}{A vqa} & 100.0&100.0&100.0&100.0&100.0&100.0&100.0&100.0&100.0&100.0&100.0&100.0$\pm$0\\
        
        \cmidrule{2-15}
        \multirow{6}{*}{\STAB{\rotatebox[origin=c]{60}{$\texttt{T-SHAP}_c$}}}
        & \multirow{1}{*}{CLIP} & 44.7&52.3&51.5&51.8&52.1&50.9&50.0&49.7&52.1&52.6&49.9&{\sethlcolor{lightgreen}\hl{50.7}}$\pm$2 & bal.\\
        & \multirow{1}{*}{LXMERT} & 51.7&{\bf37.1}&46.5&\enumber{47.3}&\enumber{46.4}&{\bf36.6}&42.1&42.2&{\bf38.2}&{\bf37.2}&\redtable{\bf{36.1}}&{\sethlcolor{lightgreen}\hl{41.9}}$\pm$5 & vis.\\
        & \multirow{1}{*}{A mscoco} & 56.7&{\bf63.5}&58.3&58.0&59.5&{\bf64.1}&{\bf61.7}&{\bf61.5}&{\bf61.9}&{\bf61.4}&{\bf63.9}&60.9$\pm$3 & txt.\\
        & \multirow{1}{*}{A flickr} & 59.5&\bf{61.7}&59.6&59.8&59.5&\bf{61.6}&59.8&58.9&60.9&{\bf61.9}&{\bf63.5}&60.6$\pm$1 & txt.\\
        & \multirow{1}{*}{A refcoco} & 53.3&57.2&55.4&\enumber{55.1}&\enumber{55.8}&57.0&54.5&54.4&57.9&58.9&\redtable{56.8}&{\sethlcolor{lightpurple}\hl{56.0}}$\pm$2 & txt.\\
        & \multirow{1}{*}{A vqa} & {\bf64.6}&{\bf63.6}&{\bf62.5}&{\bf61.4}&{\bf63.4}&{\bf63.0}&59.3&60.3&{\bf63.6}&{\bf63.1}&{\bf62.1}&{\bf{\sethlcolor{lightgreen}\hl{62.4}}}$\pm$2 & txt.\\
        
        \cmidrule{2-15}
        \multirow{6}{*}{\STAB{\rotatebox[origin=c]{60}{$\texttt{T-SHAP}_f$}}}
        & \multirow{1}{*}{CLIP} & 45.2&53.0&50.8&51.7&51.1&51.0&48.3&48.2&52.4&52.1&50.0&50.3$\pm$2 & bal.\\
        & \multirow{1}{*}{LXMERT} & 52.3&{\bf39.4}&48.2&48.8&45.8&{\bf36.5}&43.9&42.7&{\bf39.1}&{\bf38.6}&\redtable{45.0}&43.7$\pm$5& vis.\\
        & \multirow{1}{*}{A mscoco} & 57.2&{\bf62.8}&57.7&56.0&57.0&{\bf64.6}&{\bf61.9}&{\bf63.2}&{\bf61.9}&{\bf61.8}&{\bf65.8}&60.9$\pm$3 & txt.\\
        & \multirow{1}{*}{A flickr} & 56.1&{\bf61.9}&57.8&57.8&58.5&\bf{62.5}&59.3&{\bf61.9}&{\bf61.1}&{\bf62.1}&{\bf61.7}&60.1$\pm$2 & txt.\\
        & \multirow{1}{*}{A refcoco} & 56.1&58.5&56.2&55.6&57.8&57.6&55.5&56.9&58.4&58.4&\redtable{\bf{61.3}}&57.5$\pm$2 & txt.\\
        & \multirow{1}{*}{A vqa} & {\bf64.0}&{\bf64.7}&{\bf61.9}&{\bf60.9}&{\bf61.2}&{\bf63.2}&59.9&60.1&{\bf63.4}&{\bf62.4}&{\bf62.2}&{\bf62.2}$\pm$2 & txt.\\
        
        \bottomrule
    \end{tabular}    }
    \caption{ Performance and multimodal score of VL models on the instruments of the \protect\VALSE{} benchmark. We bold-face high accuracies and multimodally unbalanced models on tasks.
    $acc_r$ is the pairwise ranking accuracy,
    considering predictions as correct if $p(caption,img) > p(foil,img)$.
    Overall foil task performance $acc$ is the mean of $acc_c$ and $acc_f$ (equal number of samples, all pairs).
    {\bf A} stands for ALBEF models fine-tuned on different tasks and datasets: image retrieval on MSCOCO and Flickr30k, visual grounding on RefCOCO+ and VQA.
    $\dagger${\bf bal.} Counting balanced. $\dagger${\bf sns.} Counting small numbers. {\bf adv.} Counting adversarial. {\bf repl.} Action replacement. {\bf swap.} Actant swap. $\ddagger$ {\bf Sp.rel.} Spatial relations. $\dagger${\bf std.} Coreference standard. {\bf MM skew}: Modality on which a model relies more: {\bf bal.} balanced, {\bf vis.} visual, {\bf txt.} textual.    We test CLIP in pairwise ranking mode only (CLIP works contrastively).    }
    \label{tab:valse-results-appendix}
\end{table*}

\section{Additional results}\label{app:detailed-results}
Due to space constraints, we could not include full detailed results on \VALSE{} in \ref{tab:valse-results-main}. Here, we present Table \ref{tab:valse-results-appendix}, which is an extended version of Table \ref{tab:valse-results-main} including the MM-SHAP scores for foils too, rather than just the captions. It also includes fanned out accuracies over matching image-captions $acc_c$ and mismatching image-foils $acc_f$.

\subsection{Correlation between accuracy and MM-SHAP}\label{app:shap-acc}
For each model and instrument on \VALSE{}, we computed the Spearman correlation coefficient
between the sample's accuracy and textual degree. The correlations are very low, e.g., the correlation between $acc_c$ and $\texttt{T-SHAP}_c$ is  around 0.02 for most instruments and models, rising to 0.12 in rare cases. This low correlation between accuracy and MM-SHAP indicates that they are not measuring the same aspect: accuracy measures the models' performance while MM-SHAP measures the degree to which a modality was used -- independently of the success of its use.

\subsection{MM-SHAP difference between captions and foils}\label{app:diff-cap-foils}

We do not find notable differences between foils and captions on \VALSE{} in terms of MM-SHAP (cf. Table \ref{tab:valse-results-appendix}), while we find clear differences in accuracies between $acc_c$ and $acc_f$, since they measure the model's preference towards one side in the binary classification. Similar MM-SHAP scores between captions and foils speak for their ability to capture how the model's input matters for the prediction, independently on which class the decision falls onto. A notable exception is the difference between $\texttt{T-SHAP}_c$ and $\texttt{T-SHAP}_f$ for LXMERT and ALBEF refoco on Foil-it! (underlined numbers in Table \ref{tab:valse-results-appendix}).

\section{Sample-level Analyses with MM-SHAP} \label{app:samples}
SEE FIGURES ON FOLLOWING PAGES!

\begin{figure}[ht!]\centering
    \includegraphics[width=\linewidth]{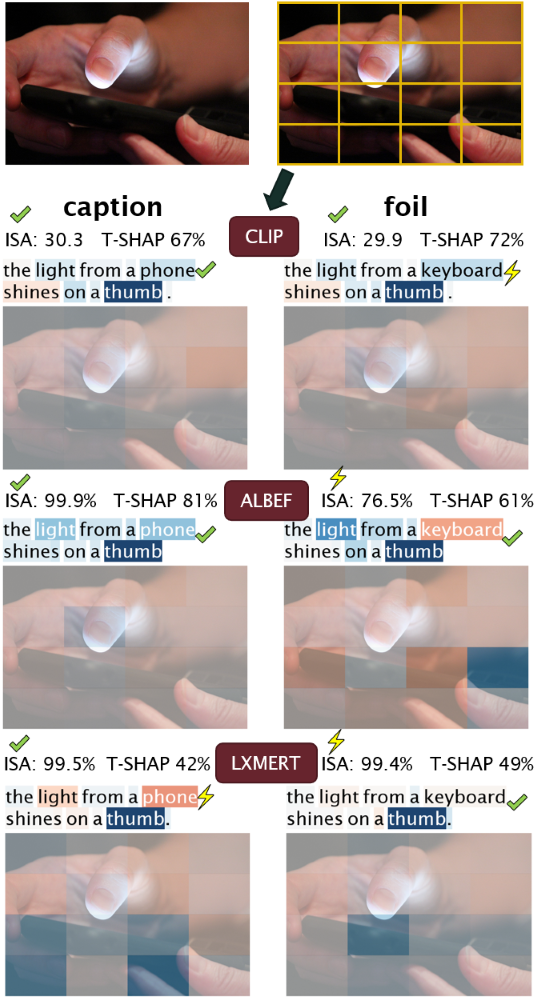}
    \caption{ \textbf{Low discrepancy} \emph{noun phrase} foil: Image-sentence alignment score (ISA) of the six VL models with their textual degree \texttt{T-SHAP} (in \%). Each text and image token (image patch) is colour-coded: Blue tokens contribute to a high ISA, while red ones lower the ISA. The visual degree is $100 - \texttt{T-SHAP}$. Note that the ISA of CLIP is an absolute score, while ALBEF and LXMERT predict ISA probabilities. With \protect\check{} we mark correct ISA and highlight the correct / foil token that contributes in the right direction for aligning the image and the caption. With \protect\blitz{}, we mark incorrect ISA and wrong contribution directions.  }
\label{fig:app-version-fig1}
\end{figure}

\begin{figure*}[ht]\centering
    \includegraphics[width=0.5\textwidth]{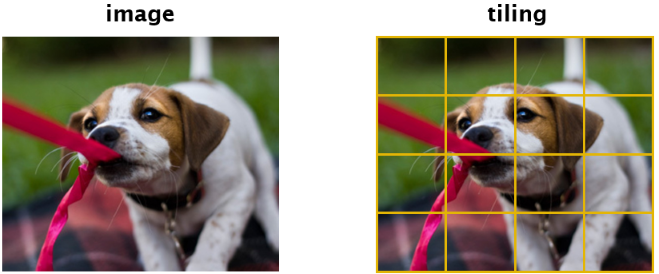}
    \includegraphics[width=\linewidth]{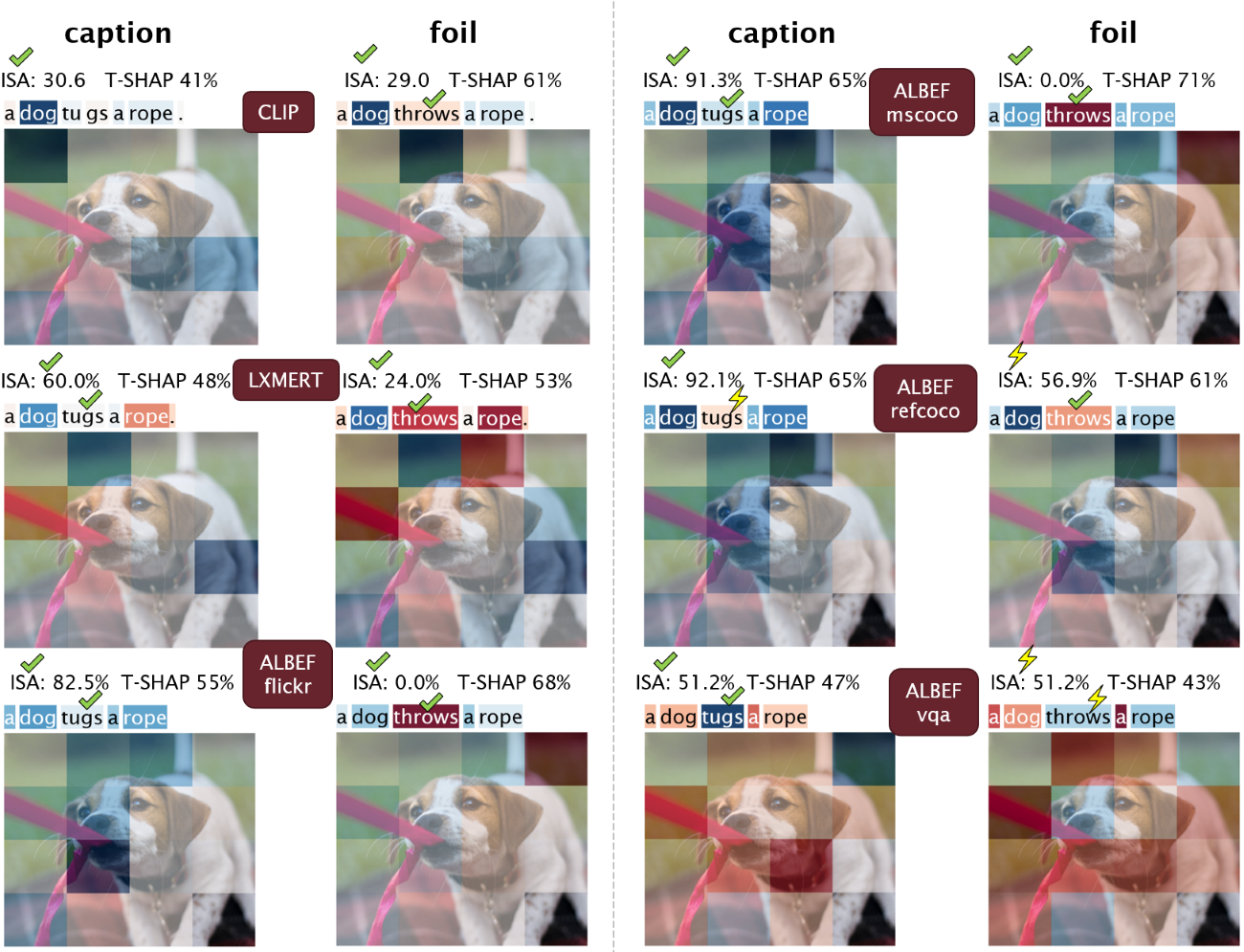}
    \caption{ \textbf{Low discrepancy} (\protect\VALSE{} \emph{action replacement}): Image-sentence alignment score (ISA) of the six VL models with their textual degree \texttt{T-SHAP} (in \%). Each text and image token (image patch) is colour-coded: Blue tokens contribute to a high ISA, while red ones lower the ISA. The visual degree is $100 - \texttt{T-SHAP}$. Note that the ISA of CLIP is an absolute score, while ALBEF and LXMERT predict ISA probabilities. With \protect\check{} we mark correct ISA and an highlight the correct / foil token that contributes in the right direction for aligning the image and the caption. With \protect\blitz{}, we mark incorrect ISA and wrong contribution directions.     }
\label{fig-app:actions-easy}
\end{figure*}

\begin{figure*}[ht]\centering
    \includegraphics[width=0.5\textwidth]{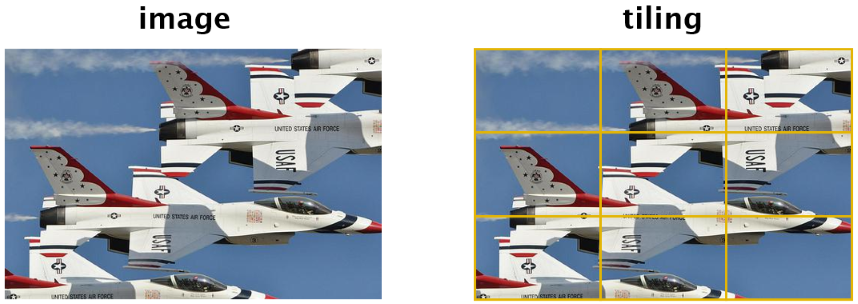}
    \includegraphics[width=\linewidth]{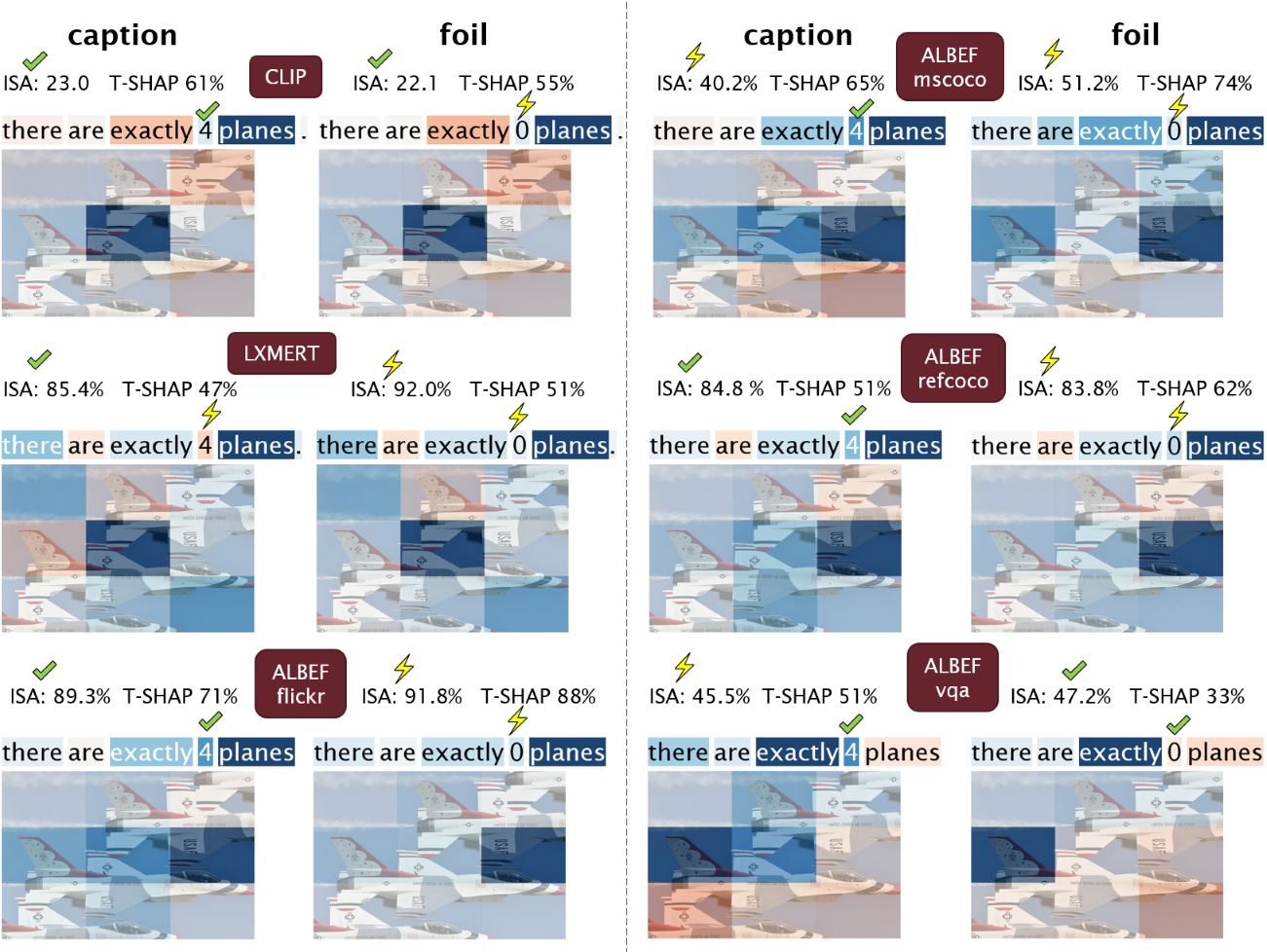}
    \caption{ \textbf{Low discrepancy} (\protect\VALSE{} \emph{counting}): Image-sentence alignment score (ISA) of the six VL models with their textual degree \texttt{T-SHAP} (in \%). Each text and image token (image patch) is colour-coded: Blue tokens contribute to a high ISA, while red ones lower the ISA. The visual degree is $100 - \texttt{T-SHAP}$. Note that the ISA of CLIP is an absolute score, while ALBEF and LXMERT predict ISA probabilities. With \protect\check{} we mark correct ISA and an highlight the correct / foil token that contributes in the right direction for aligning the image and the caption. With \protect\blitz{}, we mark incorrect ISA and wrong contribution directions.     }
\label{fig-app:counting-easy}
\end{figure*}

\begin{figure*}[ht]\centering
    \includegraphics[width=0.4\textwidth]{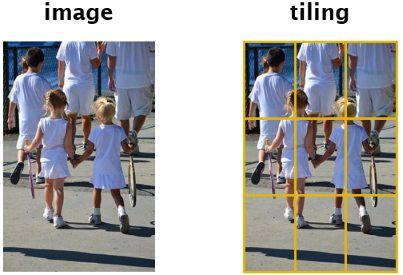}
    \includegraphics[width=\linewidth]{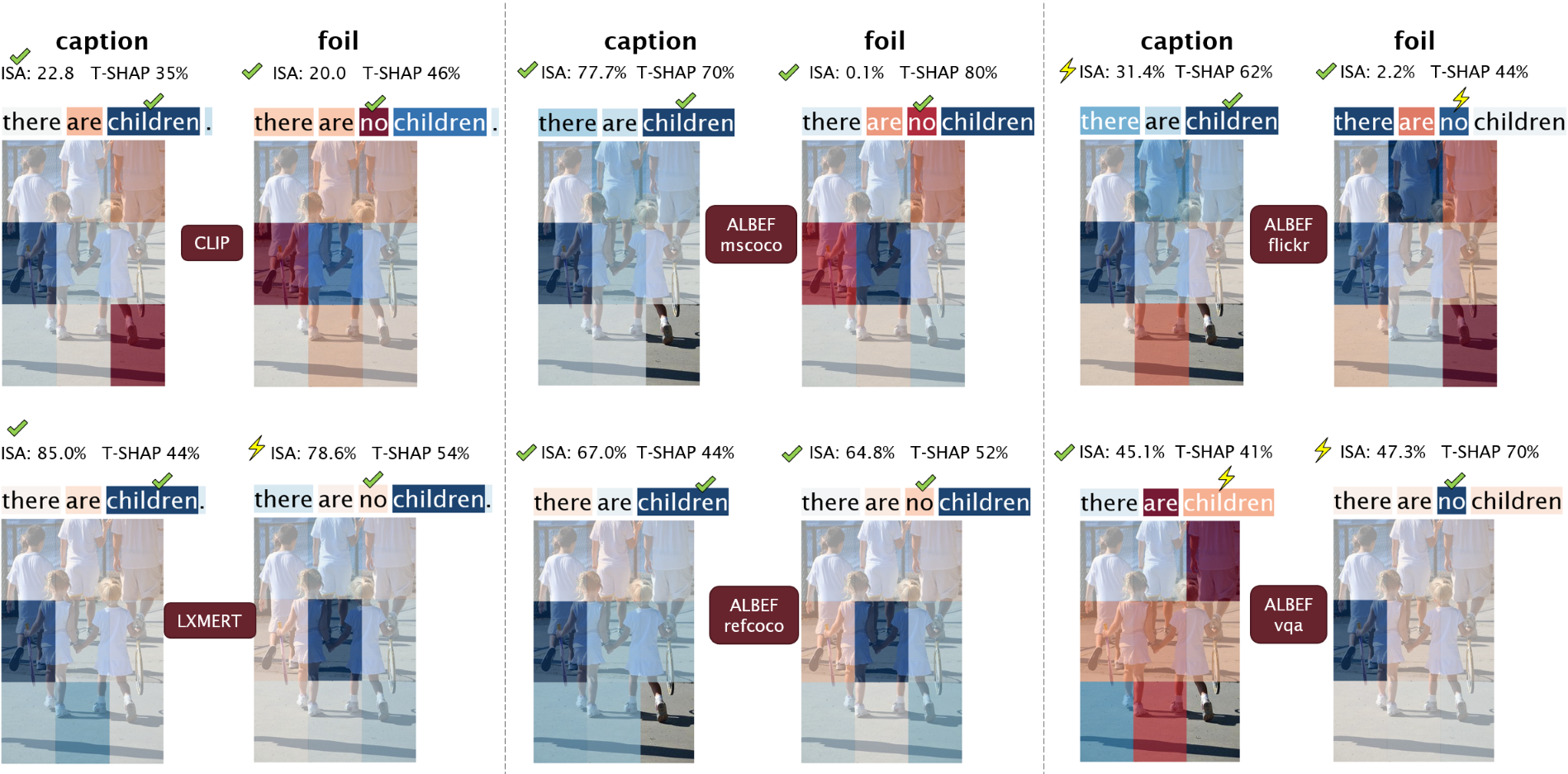}
    \caption{ \textbf{Low discrepancy} (\protect\VALSE{} \emph{existence positive}): Image-sentence alignment score (ISA) of the six VL models  with their textual degree \texttt{T-SHAP} (in \%). Each text and image token (image patch) is colour-coded: Blue tokens contribute to a high ISA, while red ones lower the ISA. The visual degree is $100 - \texttt{T-SHAP}$. Note that the ISA of CLIP is an absolute score, while ALBEF and LXMERT predict ISA probabilities. With \protect\check{} we mark correct ISA and an highlight the correct / foil token that contributes in the right direction for aligning the image and the caption. With \protect\blitz{}, we mark incorrect ISA and wrong contribution directions.     }
\label{fig-app:existence-positive}
\end{figure*}

\begin{figure*}[!ht]\centering
    \includegraphics[width=0.5\textwidth]{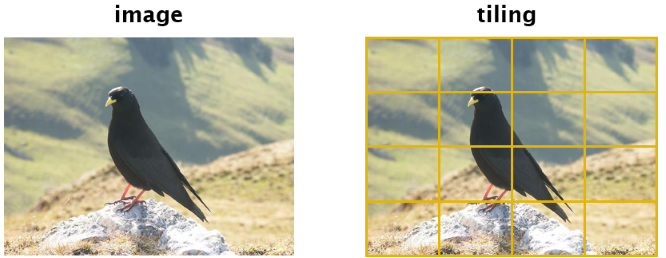}
    \includegraphics[width=\linewidth]{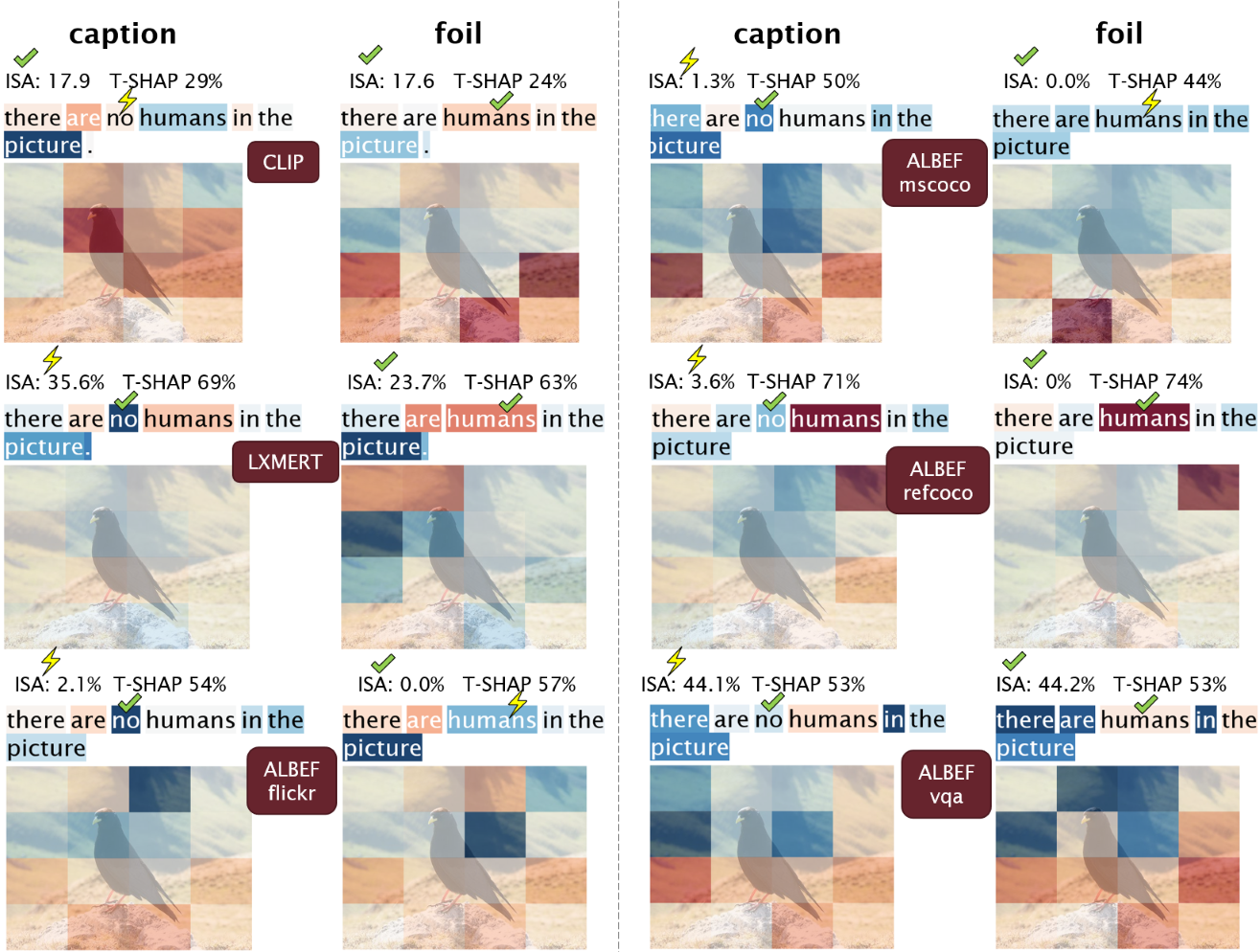}
    \caption{ \textbf{Low discrepancy} (\protect\VALSE{} \emph{existence negative} -- harder phenomenon than positive existence): Image-sentence alignment score (ISA) of the six VL models with their textual degree \texttt{T-SHAP} (in \%). Each text and image token (image patch) is colour-coded: Blue tokens contribute to a high ISA, while red ones lower the ISA. The visual degree is $100 - \texttt{T-SHAP}$. Note that the ISA of CLIP is an absolute score, while ALBEF and LXMERT predict ISA probabilities. With \protect\check{} we mark correct ISA and an highlight the correct / foil token that contributes in the right direction for aligning the image and the caption. With \protect\blitz{}, we mark incorrect ISA and wrong contribution directions.     }
\label{fig-app:existence-negative}
\end{figure*}

\begin{figure*}[!ht]\centering
    \includegraphics[width=0.5\textwidth]{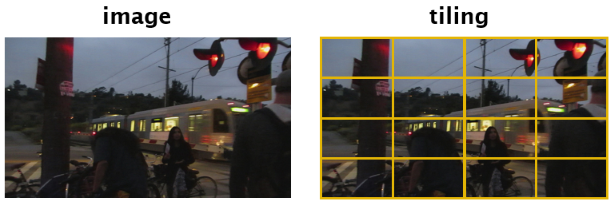}
    \includegraphics[width=\linewidth]{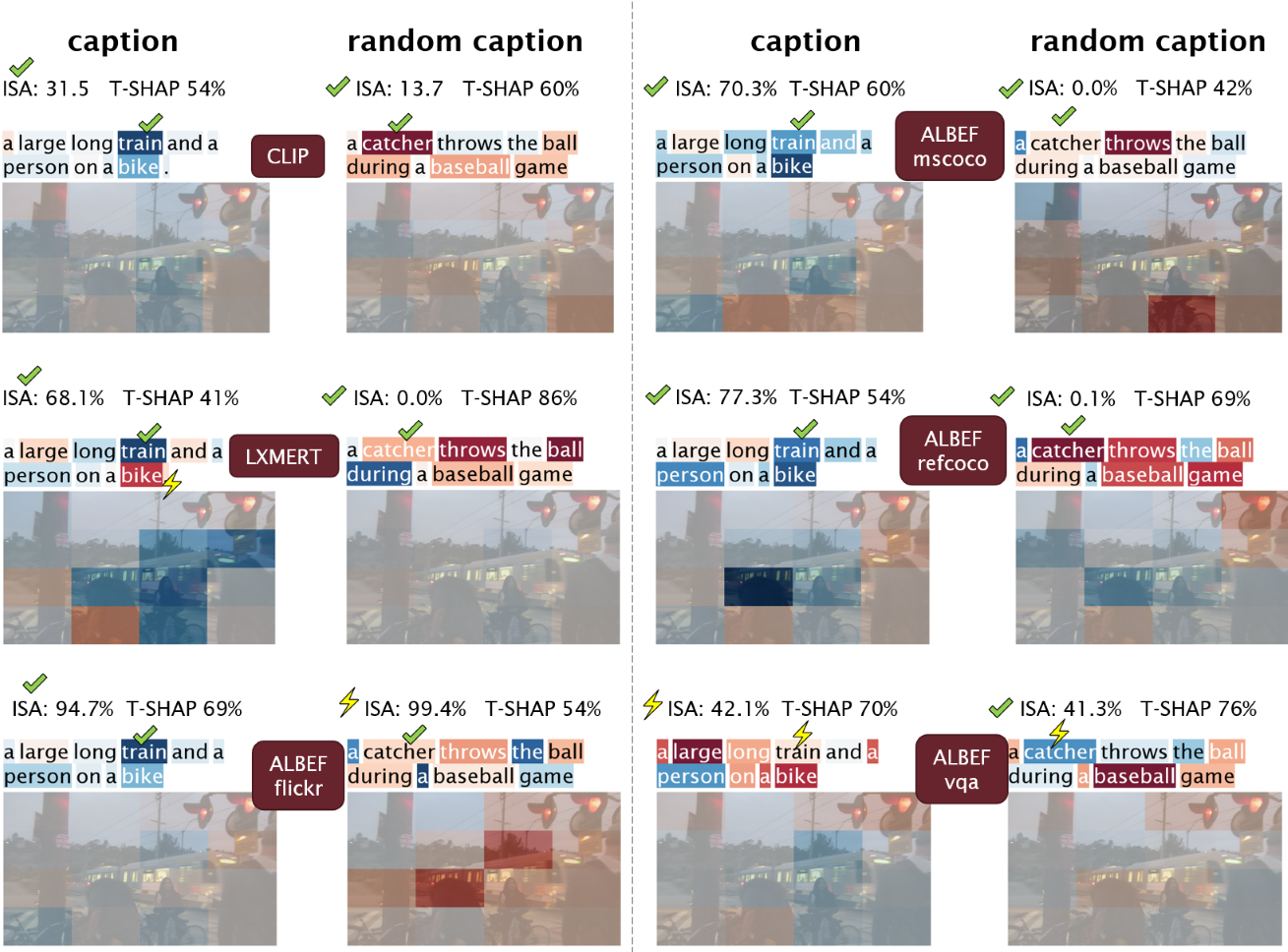}
    \caption{ \textbf{High discrepancy} (MSCOCO): Image-sentence alignment score (ISA) of the six VL models with their textual degree \texttt{T-SHAP} (in \%). Each text and image token (image patch) is colour-coded: Blue tokens contribute to a high ISA, while red ones lower the ISA. The visual degree is $100 - \texttt{T-SHAP}$. Note that the ISA of CLIP is an absolute score, while ALBEF and LXMERT predict ISA probabilities. With \protect\check{} we mark correct ISA and an highlight one important token that contributes in the right direction for aligning the image and the caption. With \protect\blitz{}, we mark incorrect ISA and wrong contribution directions.     }
\label{fig-app:high-discr-easy-train}
\end{figure*}

\begin{figure*}[!ht]\centering
    \includegraphics[width=0.3\textwidth]{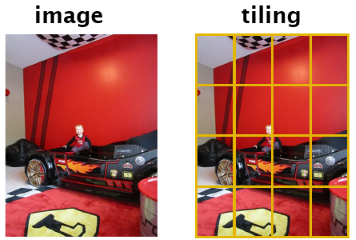}
    \includegraphics[width=\linewidth]{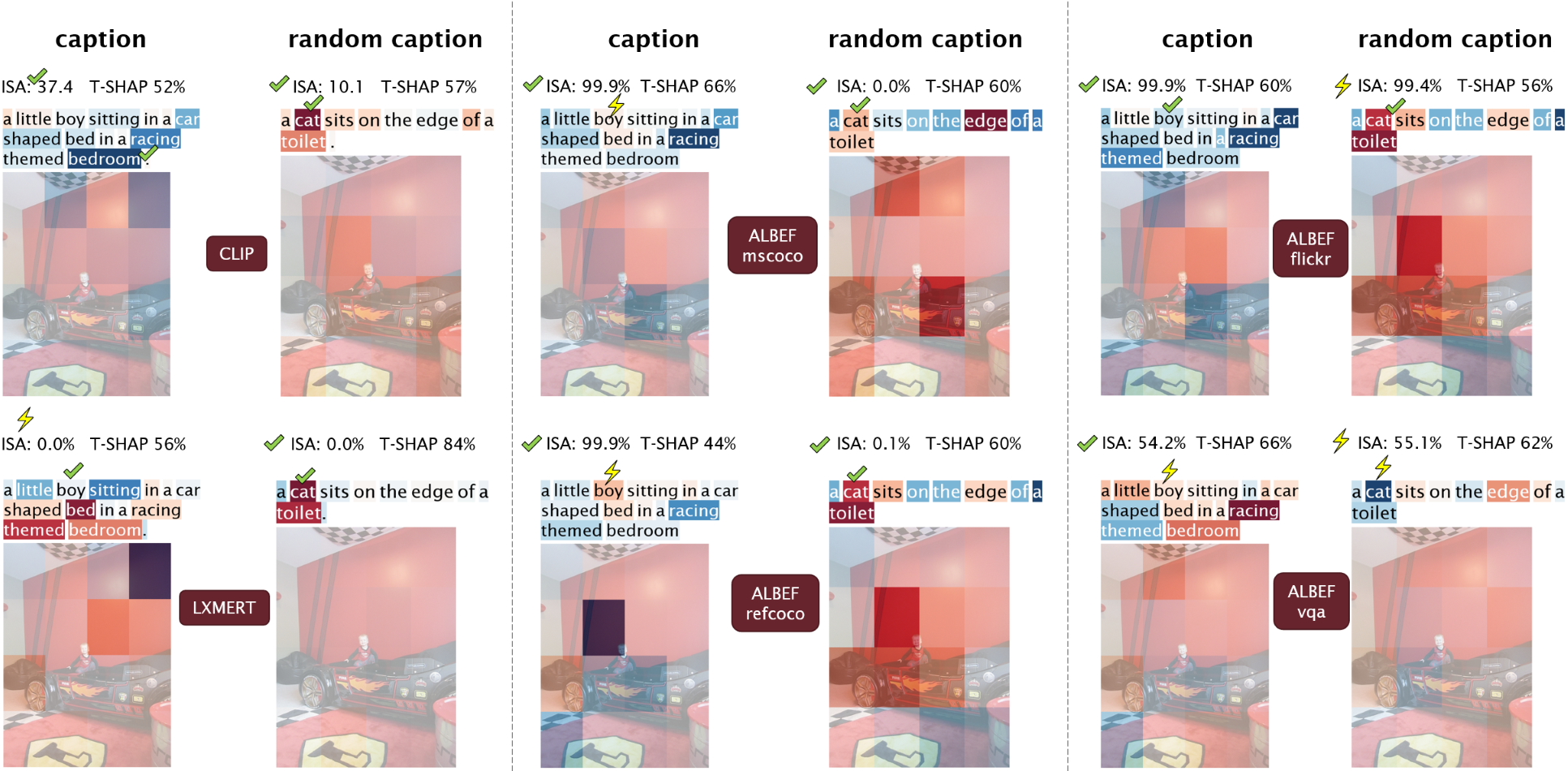}
    \caption{ \textbf{High discrepancy} (MSCOCO) \emph{hard example} where the models have trouble recognising the bed: Image-sentence alignment score (ISA) of the six VL models with their textual degree \texttt{T-SHAP} (in \%). Each text and image token (image patch) is colour-coded: Blue tokens contribute to a high ISA, while red ones lower the ISA. The visual degree is $100 - \texttt{T-SHAP}$. Note that the ISA of CLIP is an absolute score, while ALBEF and LXMERT predict ISA probabilities. With \protect\check{} we mark correct ISA and highlight one important token  that contributes in the right direction for aligning the image and the caption. With \protect\blitz{}, we mark incorrect ISA and wrong contribution directions. }    
\label{fig-app:high-discr-hard}
\end{figure*}

\begin{figure*}[!ht]\centering
    \includegraphics[width=0.5\textwidth]{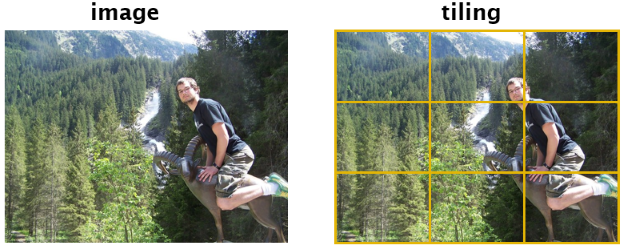}
    \includegraphics[width=\linewidth]{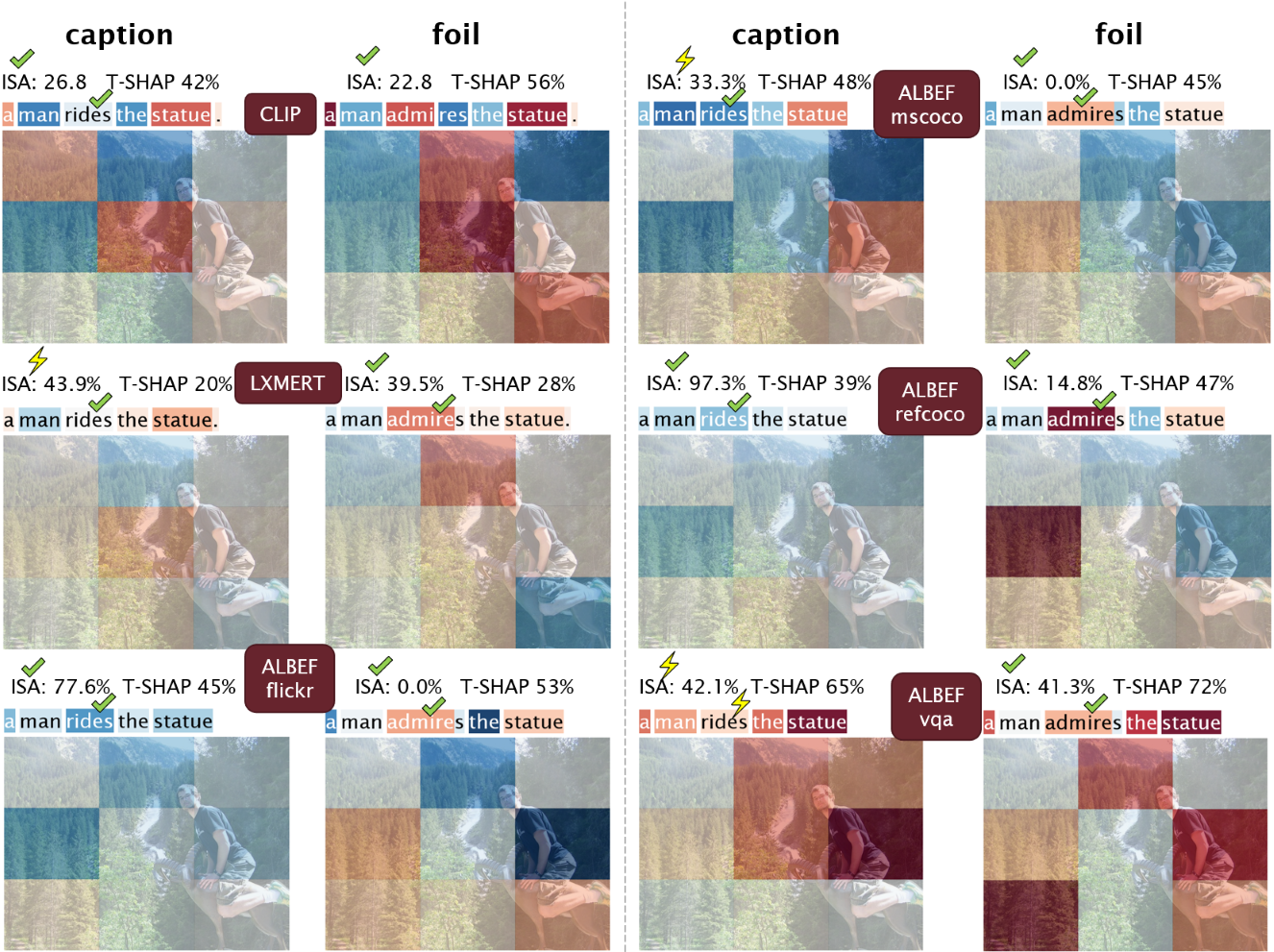}
    \caption{ \textbf{Low discrepancy} (\protect\VALSE{} \emph{action replacement}) -- \emph{hard example} where models and humans have trouble recognising the goat as a statue): Image-sentence alignment score (ISA) of the six VL models with their textual degree \texttt{T-SHAP} (in \%). Each text and image token (image patch) is colour-coded: Blue tokens contribute to a high ISA, while red ones lower the ISA. The visual degree is $100 - \texttt{T-SHAP}$. Note that the ISA of CLIP is an absolute score, while ALBEF and LXMERT predict ISA probabilities. With \protect\check{} we mark correct ISA and highlight the correct / foil token that contributes in the right direction for aligning the image and the caption. With \protect\blitz{}, we mark incorrect ISA and wrong contribution directions.   }
\label{fig-app:actions-hard}
\end{figure*}

Figures \ref{fig:app-version-fig1} to \ref{fig-app:actions-hard} contain sample-level visualisations for each model for images and i) captions that match and ii) foils / random captions that show 
low / high discrepancy mismatch with the images, as introduced in Section \ref{sec:exps-results}:
\begin{itemize}
    \item There is \textbf{low discrepancy} between images and foils obtained from \VALSE{} targeting specific linguistic phenomena, with only a phrase differing between the caption and the foil. We selected examples for different phenomena: Figure \ref{fig:app-version-fig1} (noun phrase), \ref{fig-app:actions-easy} (action replacement, easy example), \ref{fig-app:counting-easy} (counting), \ref{fig-app:existence-positive} (positive existence), \ref{fig-app:existence-negative} (negative existence), \ref{fig-app:actions-hard} (action replacement, hard example). 
    \item There is \textbf{high discrepancy} between MSCOCO images and randomly chosen captions in terms of low ISA between image and random caption -- Figures \ref{fig-app:high-discr-easy-train} (easier example) and \ref{fig-app:high-discr-hard} (harder example).
\end{itemize}


In Figure \ref{fig:app-version-fig1} we reiterate 
Figure \ref{fig:illustrated-contributions} from the main paper with more detail:
\begin{itemize}
    \item CLIP correctly predicts a foil in the pairwise accuracy setting, since the ISA score for the caption (30.3) is higher than for the foil (29.9), but fails to identify that ``keyboard'' should not contribute towards a high ISA. It successfully predicts caption alignment, but seems to misunderstand the meaning of the word ``shines'' and its instantiation in the image.
    \item ALBEF mscoco is the only model to predict ISA (99.4\%) on the caption with coherent -- but mostly textual -- indicators. It fails on foil prediction, still relying on the same textual indicators, and on the visual side \emph{focuses on counter-evidence regions}, erroneously taking them as positive support for ISA.
    \item LXMERT predicts correct ISA for the caption (99.5\% ISA), using few relevant textual tokens as indicators, and possibly useful supporting visual tokens (focuses the fingers of the two hands). It fails to detect the foil (99.4\% ISA which is higher than a 50\% classification threshold and just slightly below
    the ISA for the caption): counterevidence from textual tokens is out-weighted by a single strong indicator (thumb); visual tokens confirm ISA despite focusing on  counterevidence (the phone).
\end{itemize}

\noindent
On the following pages we present Figures \ref{fig-app:counting-easy} to \ref{fig-app:actions-hard} with more samples and their analyses.

We sampled the instances based on 
the following criteria: i) low / high discrepancy; ii) interesting \VALSE{} instruments; iii) easier (no cluttering, no dark spots, no blur) and iv) harder examples (e.g., hard to recognise the statue as such in Figure \ref{fig-app:actions-hard}).


Through Fig.\ \ref{fig-app:counting-easy} to \ref{fig-app:actions-hard}, we observe some patterns:
\paragraph{Model performance does not tell much about the multimodal degree.} A correct ISA score (high for the caption, low for the random caption/foil) is not always accompanied by a sensible contribution pattern in terms of Shapley values as seen for example in Figures \ref{fig:app-version-fig1} and \ref{fig-app:counting-easy} for CLIP and LXMERT. The Shapley values computed on the image and text side deliver much better intuition about what was successfully aligned and what was not grounded correctly. Among all models, LXMERT seems to be most affected by high discrepancy between performance and image and text token contributions.

\paragraph{Easy examples deliver more robust contribution patterns.} On easy examples (Figures \ref{fig-app:actions-easy} and \ref{fig-app:counting-easy}), where the model generally performs well, we can see how in the low discrepancy cases where caption and foil differ in only one word, the one word difference does not change the contribution patterns much. In contrast, low discrepancy hard examples (Figures \ref{fig-app:high-discr-hard} -- unusual bed and bedroom arrangement and \ref{fig-app:actions-hard} -- hard to recognise the goat as a statue without world knowledge) deliver different patterns on caption and foil, indicating confusion from the models.

\paragraph{Positive existence is easier than negative existence.}
When comparing Figures \ref{fig-app:existence-positive} and \ref{fig-app:existence-negative} we get some insight into how the models' image-sentence alignment pretraining objective affects their behaviour:

For positive existence, where the caption indicates that \textbf{an object is present in the image} -- as  in Fig.\ \ref{fig-app:existence-positive}: \emph{There are children.} -- is better handled by the models, delivering more sensible patterns for image-caption pairs. The contribution patterns on the negated version of the existence sentence -- the foil \emph{There are no children.} -- show that some models handled the negation correctly (CLIP, LXMERT, ALBEF mscoco and refcoco), while the rest do not.

Negative existence, where the caption indicates that \textbf{an object is not present in the image} -- as seen in Fig.\ \ref{fig-app:existence-negative}: \emph{There are no humans in the picture.} -- seems more difficult to align, since the objects are 
not present in the image and to assign a high ISA for text mentions that cannot be located, the model needs to understand the negation.
The foil, changing the sentence to affirmative -- \emph{There are humans in the picture.} -- turns the instance into a much simpler case of no image-sentence alignment, as is often seen during pretraining. Unsurprisingly, all models correctly predict a low ISA in Figure \ref{fig-app:existence-negative}.

\paragraph{Counting is hard.}
In Figure \ref{fig-app:counting-easy} for the counting foils in \VALSE{}, CLIP is the only model that assigns higher ISA for the image-caption pair and not to the image-foil pair. Overall, the contribution patterns look scattered: High visual contributions in the image indicate that the models align the plane object to its mention in the sentence, but we see
confused textual contributions from the mentioned number of planes (0 or 4) in the text. This is unsurprising, given the low performance of VL models in counting as highlighted by \citet{parcalabescu-etal-2021-seeing}.

\section{Why not to use Attention for defining a Multimodality Score}\label{app:against-attention}

\subsection{Requirements for a MM Score}\label{app:requirements}
For defining a multimodality score that aims at quantifying each modality's contribution to any model prediction, we need an interpretability method that has crucial properties to do so.
With the properties of efficiency, symmetry, dummy variable, additivity (see §\ref{sec:shap-backgr}), Shapley values provide important ingredients
for \emph{sample-based explanations} that can be aggregated in a straightforward way into \emph{dataset-level explanations} for machine learning methods \cite{covert2020SAGE}. Other interpretability methods lack the robustness and theoretical foundation to produce a  multimodality score that is comparable to the one proposed in our work.

In particular, attention -- while being widely used for generating visually appealing heat-maps -- does not fulfil the condition of delivering a fair payout (like Shapley values do) and it is questionable how much high/low attention scores correlate with high/low contributions of input features for system predictions \cite{jain-wallace-2019-attention, wiegreffe-pinter-2019-attention}.\footnote{Arguably this may be the case when attention weights are high, but it is clearly not the case 
when attention weights are low.} Attention linearly combines input features and determines how much of each token is mixed with every other token. But it does not necessarily mean that a low attention value cannot have a large impact on the decision of the model. In other words, a pinch of salt
is enough to make food taste good: 
Even if the attention score for salt is low, its contribution to the taste of the food (captured by Shapley values) is high.

Attention is present in transformers in multiple layers and to complicate the matter even further, each attention layer contains multiple attention heads. Hence, to visualise attention we need a carefully designed interface, as proposed, e.g., by \citet{jaunet2021visqa} \url{https://visqa.liris.cnrs.fr/} to keep a reasonable overview of all attention values. When integrating the multiple attention values and propagating them back to the input to assign relevancy values for image and text tokens, research strives to generate simple explanations that represent the most important tokens and tend to inhibit the rest, as can be seen on the progress from \citet{chefer2021transformer} to \citet{chefer2021generic} (cf.\ Figure 4 in \citet{chefer2021generic}).

\subsection{Measuring negative contribution}\label{app:attention-no-neg}
While Shapley values estimate both the positive and the \emph{negative contributions} of input tokens towards the model prediction -- which is relevant for foil words --, attention \cite{chefer2021generic} allows for positive-only relevance assessments.

\begin{figure*}[t]\centering
    \includegraphics[width=0.7\textwidth]{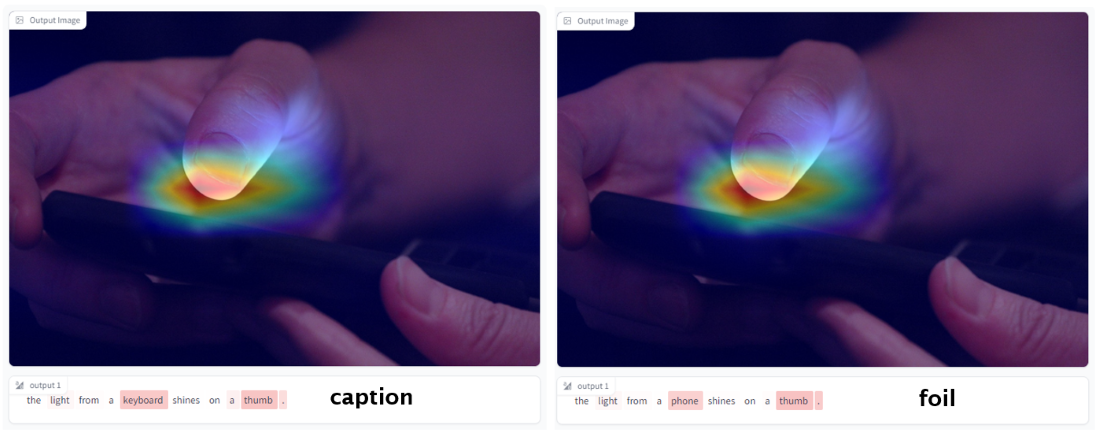}
    \includegraphics[width=0.7\textwidth]{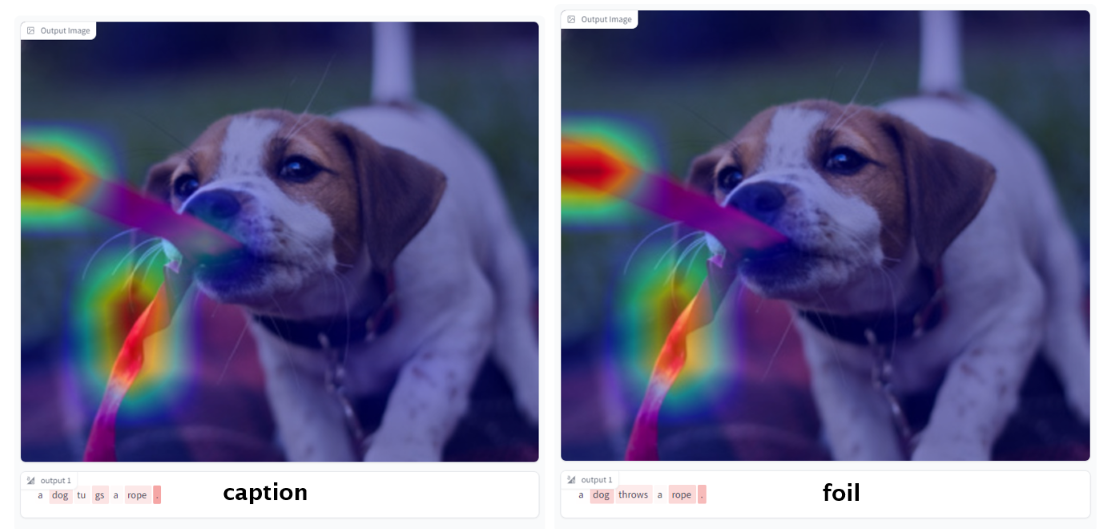}
    \includegraphics[width=0.7\textwidth]{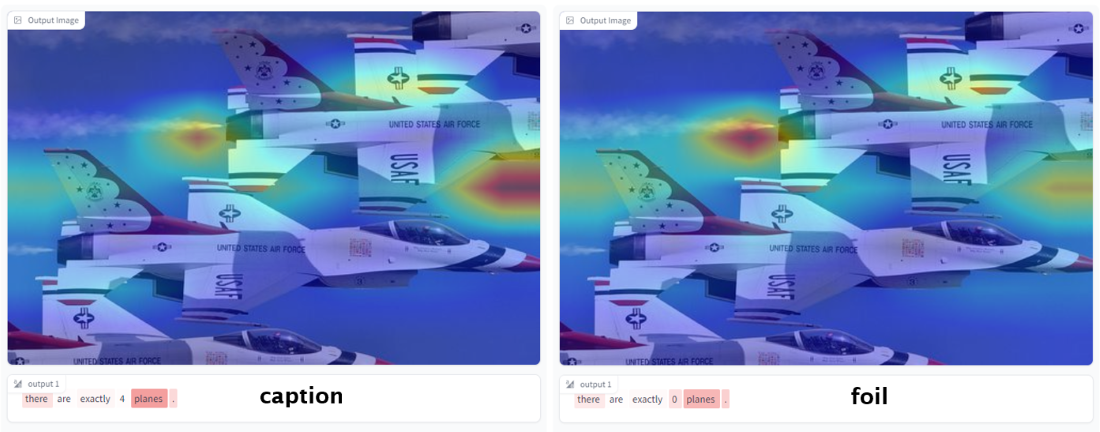}
    \includegraphics[width=0.5\textwidth]{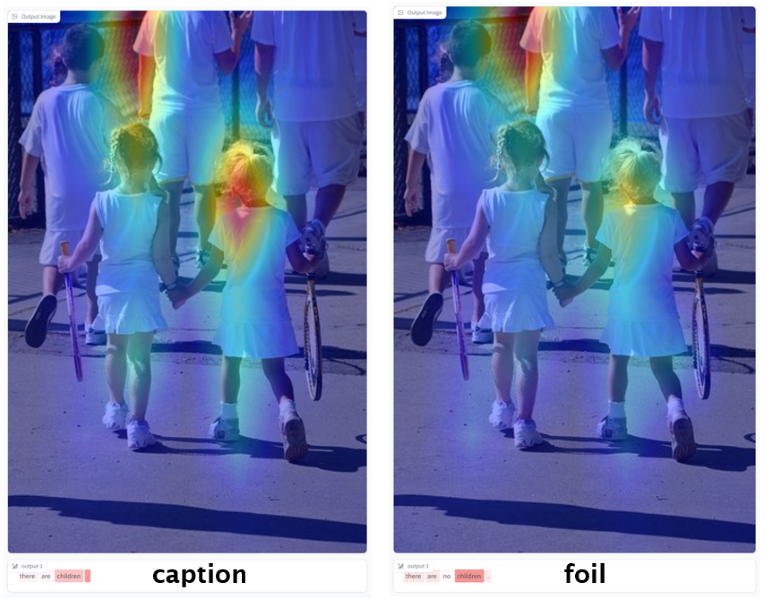}
    \caption{ \textbf{Low discrepancy}. CLIP results of attention-based relevance visualisation, using the method of \citet{chefer2021generic} \url{https://huggingface.co/spaces/PaulHilders/CLIPGroundingExplainability}. Red means high relevancy, blue is zero relevancy and there is no negative relevancy (while Shapley values do allow for positive and negative contributions). Note that the heat-maps give the impression that the relevance irradiates from single spots. This is an artefact from the visualisation since the model works with 32x32 pixel patches and it is these patches that each have a relevance score. For reference: the images are around 500 pixels in height and width.    }
\label{fig-app:attention-low-discr}
\end{figure*}

\begin{figure*}[t]\centering
    \includegraphics[width=0.9\textwidth]{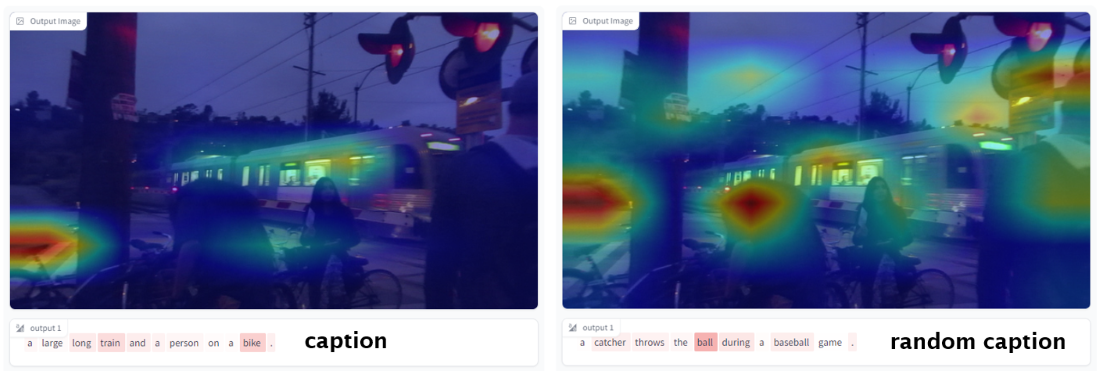}
    \caption{ \textbf{High discrepancy}. CLIP results of attention-based relevance visualisation, using the method of \citet{chefer2021generic} \url{https://huggingface.co/spaces/PaulHilders/CLIPGroundingExplainability}. Red means high relevancy, blue is zero relevancy and there is no negative relevancy (while Shapley values do allow for positive and negative contributions). Note that the heat-maps give the impression that the relevance irradiates from single spots. This is an artefact from the visualisation since the model works with 32x32 pixel patches and it is these patches that each have a relevance score. For reference: the images are around 500 pixels in height and width.     }
\label{fig-app:attention-high-discr}
\end{figure*}

In Figures \ref{fig-app:attention-low-discr} and \ref{fig-app:attention-high-discr}, we have visualised CLIPs attention-based relevancy for the image-caption and foil examples shown in Figures \ref{fig:app-version-fig1} to \ref{fig-app:high-discr-easy-train} using the method of \citet{chefer2021generic}. On the image side, we observe little to no changes in the attention visualisation, when comparing image-caption to image-foil pairs (cf. Figure \ref{fig-app:attention-low-discr}). Even more, on the text side, both the correct and the foil word carry relatively similar attention scores, with no indication whether this contributes positively or negatively towards the model prediction. Shapley values however, are sensitive to foil words and we can visualise whether the word contributes towards raising the ISA (high image-sentence match) or lowering the ISA (e.g., Figure \ref{fig-app:actions-easy}).

Besides the problematic interpretation of attention as feature contribution and the many ways of aggregating and propagating the different attention values to the input, another problem with attention is that it is unclear how to disentangle and aggregate the textual self-attention, visual self-attention, text-to-image attention and image-to-text attention into a single multimodality score that assesses the degree to which a given modality contributes towards the model prediction.

All things considered, we argue that attention is not well-suited as a basis for a multimodality score we aim for in this work, but that Shapley values -- as presented in this paper -- are, thanks to their theoretical properties (efficiency, symmetry, dummy variable, additivity) and their property of being model-agnostic measurements of input feature contributions.

SEE FIGURES ON FOLLOWING PAGES!

\end{document}